\documentclass[11pt]{article}

\usepackage{ACL2023}
\usepackage{CJK}
\usepackage{multirow} 
\usepackage{soul} 
\usepackage{color, xcolor}
\usepackage{times}
\usepackage{latexsym}
\usepackage{amsthm}
\usepackage[T1]{fontenc}
\usepackage{graphicx}
\usepackage{float}
\usepackage{booktabs}
\usepackage{pifont} 
\usepackage[utf8]{inputenc}
\usepackage{amsmath}
\usepackage{microtype}

\usepackage{inconsolata}
\usepackage{amssymb}
\usepackage{graphicx} 
\usepackage{subcaption} 
\usepackage{cleveref}
\usepackage{enumerate}
\usepackage{array}
\usepackage{colortbl}
\usepackage{bbding}
\usepackage{tikz}
\usepackage{graphicx}
\usepackage{subcaption}
\usepackage{hyperref}
\usepackage{url}
\usepackage{booktabs}
\usepackage{tabularx}
\usepackage{wrapfig}
\usepackage{enumitem}
\usepackage{amsmath}
\usepackage{amsfonts}
\usepackage{bm}
\usepackage{bbold}
\usepackage{graphicx}
\usepackage{enumerate}
\usepackage{caption}
\usepackage{graphicx}
\usepackage{multirow}
\usepackage{xcolor}
\usepackage{subcaption}
\usepackage{booktabs}
\usepackage{colortbl}
\usepackage{algorithm}
\usepackage{caption}
\definecolor{goldenyellow}{rgb}{1.0, 0.93, 0.7}
\definecolor{deepblue}{rgb}{0.0, 0.25, 0.5}
\usepackage{xcolor}
\usepackage{soul}

\newcommand{\highlight}[1]{\colorbox{pink}{#1}}

\title{Holistic Analysis of Hallucination in GPT-4V(ision): \\Bias and Interference Challenges}

\author{Chenhang Cui$^{1*}$, Yiyang Zhou$^{1*}$, Xinyu Yang$^{2}$, Shirley Wu$^{3}$\\ \textbf{Linjun Zhang$^{4}$, James Zou$^{3}$, Huaxiu Yao$^{1}$}\\
  $^{1}$UNC-Chapel Hill, $^{2}$Carnegie Mellon University, $^{3}$Stanford University, $^{4}$Rutgers University\\
  \texttt{chenhangcui@gmail.com}, \texttt{zhouyiyangailab@gmail.com}, \texttt{huaxiu@cs.unc.edu}
}

\begin{document}
\maketitle
\def\thefootnote{$*$}\footnotetext{Equal contribution. Work was done during Chenhang Cui and Yiyang Zhou's remote internship at UNC.}

\begin{abstract}

While GPT-4V(ision) impressively models both visual and textual information simultaneously, it's hallucination behavior has not been systematically assessed. To bridge this gap, we introduce a new benchmark, namely, the \textbf{B}ias and \textbf{In}terference Challen\textbf{g}es in Visual Language M\textbf{o}dels (Bingo). This benchmark is designed to evaluate and shed light on the two common types of hallucinations in visual language models: bias and interference. Here, bias refers to the model's tendency to hallucinate certain types of responses, possibly due to imbalance in its training data. Interference pertains to scenarios where the judgment of GPT-4V(ision) can be disrupted due to how the text prompt is phrased or how the input image is presented. We identify a notable regional bias, whereby GPT-4V(ision) is better at interpreting Western images or images with English writing compared to images from other countries or containing text in other languages. Moreover, GPT-4V(ision) is vulnerable to leading questions and is often confused when interpreting multiple images together. Popular mitigation approaches, such as self-correction and chain-of-thought reasoning, are not effective in resolving these challenges. We also identified similar biases and interference vulnerabilities with LLaVA and Bard. Our results characterize the hallucination challenges in GPT-4V(ision) and state-of-the-art visual-language models, and highlight the need for new solutions. The Bingo benchmark is available at \url{https://github.com/gzcch/Bingo}.

\textbf{Caution}: This paper may contain model outputs that exhibit biases.





\end{abstract}



\section{Introduction}

\begin{figure*}
  \centering
    \begin{center}
           \includegraphics[width=\textwidth]{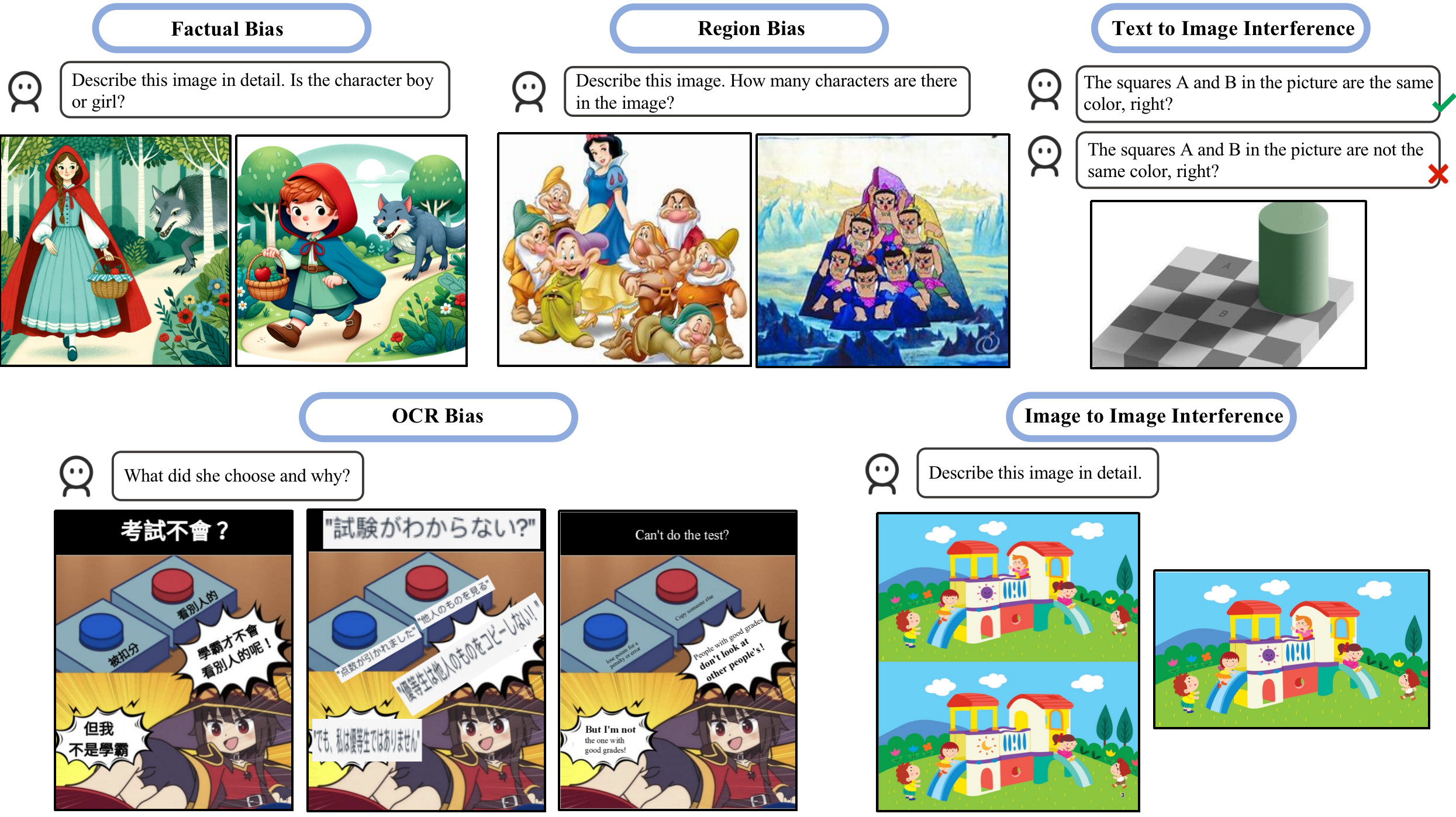}
           \captionof{figure}{
the \textbf{B}ias and \textbf{In}terference Challen\textbf{g}es in Visual Language M\textbf{o}dels (Bingo): A benchmark for a comprehensive analysis of hallucination in Vision-language Models (VLMs), including evaluations for three types of biases: factual bias, region bias and OCR bias, and two types of input interference: image-to-image interference and text-to-image interference. Here, we present a representative example for each type with both text and image inputs. 
For the demonstrated examples, each specific response from GPT-4V(ision) can be found in the appendix \ref{samples}.}
           \label{fig:BIg_case}
        \end{center}
        \vspace{-1.5em}
\end{figure*}
Large language models (LLMs), notably the GPT series developed by OpenAI, have consistently showcased remarkable capabilities spanning diverse domains \cite{radford2019language, brown2020language}. The recent release of GPT-4V(ision) further unleashes the power of connecting vision and language modalities \cite{openai2023, openai2023gpt4, openai2023gpt4vision}, capturing the attention of a wide range of researchers due to its exceptional visual capabilities across various visual comprehension and reasoning tasks~\cite{yang2023dawn}. 

However, GPT-4V(ision) may also exhibit limitations similar to other VLMs, such as LLaVA~\cite{liu2023visual}, which can easily produce hallucinations or generate inconsistent responses when presented with input images~\cite{liu2023hallusionbench, liu2023aligning, zhou2023analyzing, li2023evaluating}. In order to investigate the limitations of GPT-4V(ision) and identify situations in which it is prone to hallucinations, we construct a benchmark comprising a collection of 190 failure instances in GPT-4V(ision). Based on our observations, we have categorized these failure cases by causes of limitations in GPT-4V(ision) into bias and interference and named our benchmark as Bingo (the \textbf{B}ias and \textbf{In}terference Challen\textbf{g}es in Visual Language M\textbf{o}dels). Details are illustrated in Figure~\ref{fig:BIg_case} and described as follows.

\noindent \textbf{Bias.} Bias in GPT-4V(ision) refers to its susceptibility to generating hallucinatory outputs on specific types of examples. In Bingo, we investigate three main categories of bias, including region bias, Optical Character Recognition (OCR) bias, and factual bias. Region bias pertains to GPT-4V(ision)'s tendency to generate content biased towards specific geographic regions. OCR bias is associated with biases introduced due to limitations in OCR detectors, resulting in bias towards certain languages. Factual bias arises from the model's inclination to excessively rely on learned factual knowledge while disregarding the input image when generating responses.

\noindent \textbf{Interference.} Interference refers to scenarios in which the judgment of GPT-4V(ision) can be disrupted, making it more susceptible to hallucination. In Bingo, we conduct specific investigations into two types of interference: image-to-image interference and text-to-image interference. Image-to-image interference underscores the challenge GPT-4V(ision) faces when interpreting multiple similar images together. Text-to-image interference describes the scenarios where the human user's claims made in the text prompt can disrupt GPT-4V(ision)'s recognition capabilities.

In addition to identifying instances where GPT-4V(ision) exhibits hallucinations due to biases and interference, we have conducted a comprehensive investigation aimed at enhancing its accuracy in such scenarios. Our investigation centers on two key approaches: self-correction \cite{huang2023large} and chain-of-thoughts (CoT) reasoning \cite{wei2022chain}. In the self-correction approach, when we use the prompt "Your answer is wrong. Review your previous answer and find problems with your answer. Answer me again." after receiving an erroneous initial response, we observe a reduction in hallucinations by 16.56\%. On the contrary, in CoT reasoning, even when we employ the prompt "Let’s think step by step," we have noticed that GPT-4V(ision) still tends to produce hallucinatory responses in most instances.

To summarize, our primary contribution is curating a new benchmark to analyze the vision limitations and hallucinations of GPT-4V(ision). Our empirical analysis reveals two primary causes of GPT-4V(ision)'s hallucinations: bias and interference. We also investigate the potential solutions for rectifying these hallucinations using self-correction or chain-of-thoughts reasoning.

\section{Bingo Benchmark}

In this section, we describe our design of the Bingo benchmark. Specifically, Bingo includes 190 failure instances, along with 131 success instances as a comparison. Each image in Bingo is paired with one or two questions. Based on our observations (see details in Section~\ref{sec:analysis}), we categorize these failure cases into two categories based on the cause of hallucinations: "Interference" and "Bias". The Bias category is further divided into three types: Region Bias, OCR Bias, and Factual Bias. The Interference category is further divided into two types: Image-to-Image Interference and Text-to-Image Interference. In Table~\ref{BIg_framework}, we detail the statistics of the Bingo benchmark. We provide representative examples of each category in Figure~\ref{fig:BIg_case}.

\subsection{Bias} In Bingo, to analyze the bias in GPT-4V(ision), we collect a diverse set of images, which includes images from different regions, multilingual text within images, and images depicting content that contradicts factual knowledge. The details of this data collection are provided below:

\paragraph{Region Bias} To evaluate Region bias, we have collected data pertaining to culture, cuisine, and various other aspects from five distinct geographical regions: East Asia, South Asia, South America, Africa, and the Western world. During the data collection process, we also aim to ensure a balanced representation of image types across these regions. For example, when gathering images related to animations, we strive to match the quantity of such images for each region, thereby creating a consistent set. As shown in Figure \ref{fig:BIg_case}, we illustrate a case where we present GPT-4V(ision) with identical questions regarding animations from different regions, such as Snow White from the West and Calabash Brothers from China.

\begin{table}[t]
  \centering
  \small
\caption{Table 1 outlines the Bingo benchmark. We list the number of images and questions for each category.}
\resizebox{\linewidth}{!}{
\setlength{\tabcolsep}{1.5mm}{
\begin{tabular}{l|l|c|c}
\toprule
\multicolumn{2}{c|}{Category}  & \# Images & \# Questions \\ \midrule
\multirow{4}{*}{Bias}     & Region            & 105             & 105                \\ \cmidrule{2-4}
& OCR           & 50             & 79                \\ \cmidrule{2-4}
& Factual & 64             & 64                \\ \midrule
\multirow{2.5}{*}{Interference} & Image-to-Image  & 56             & 56                \\ \cmidrule{2-4}
 & Text-to-Image   & 33             & 66                \\
\bottomrule
\end{tabular}}}
\label{BIg_framework}
\vspace{-1em}
\end{table}

\paragraph{OCR Bias} To analyze OCR bias, we collect examples that involved obtaining images containing text within them. Subsequently, we translated this text into multiple languages, which include Arabic, Chinese, French, Japanese, and English. For example, Figure \ref{fig:BIg_case} presents a case of OCR bias, where we give GPT-4V(ision) the same question about a comic embedded with text from different countries to test its OCR capabilities in multilingual scenarios.

\paragraph{Factual Bias} To investigate whether the model excessively relies on pre-learned factual knowledge at the expense of the factual information presented in input images, we curated a set of counterfactual images. For instance, consider the factual bias case illustrated in Figure \ref{fig:BIg_case}, depicting the story of "Little Red Riding Hood." We deliberately crafted counterfactual versions of this story by substituting the girl with a boy, aiming to assess whether the model would generate responses based on its prior knowledge (i.e., that Little Red Riding Hood is traditionally portrayed as a young girl) rather than recognizing the altered fact conveyed in the image (depicting a young boy)."



\subsection{Interference}
To analyze the interference in GPT-4V(ision), we have introduced two categories of images and corresponding questions. These include interference stemming from the composition of similar images and interference arising from human users' claims within the text prompts. The specifics of this analysis are elaborated as follows:

\paragraph{Image-to-Image Interference} In image-to-image interference, we aim to determine whether GPT-4V(ision) can discern differences when presented with a set of closely resembling images. To achieve this, we curate a collection of images, each composed of several similar images. The collection includes both natural and synthetic images, with the latter primarily sourced from puzzles. Additionally, we have extracted individual images from these compositions for comparison. 
Figure \ref{fig:BIg_case} shows an example where we pieced together two slightly different, similar images to create an image-to-image interference version, and for comparison, we also included the non-composite images.

\paragraph{Text-to-Image Interference} In text-to-image interference, we aim to investigate whether GPT-4V(ision) can be influenced by human claims presented in the text prompts. To accomplish this, we curate a collection of images, each accompanied by a pair of questions. One question prompts a correct response, while the other prompts an incorrect response. For instance, as shown in the example of Figure \ref{fig:BIg_case}, when presented with an image that has two squares, A and B, of the same color, we pose two questions: "The squares A and B in the picture are the same color, right?" and "The squares A and B in the picture are not the same color, right?".

\section{Empirical Analysis}
\label{sec:analysis}
After designing the Bingo benchmark, in this section, we conduct an empirical analysis to quantify the performance of GPT-4V(ision) on Bingo benchmark in October 2023. In our analysis, we used human annotators to evaluate the accuracy of GPT-4V(ision)'s responses, assigning a score of 1 for correct answers and 0 for incorrect ones. In the remaining of this section, we will introduce our analysis of bias and interference. Moreover, we also evaluate the performance on other VLMs, such as LLaVA-1.5 and Bard.

\subsection{Analysis of Bias}
\begin{figure}[h]
  \centering
    \includegraphics[width=0.8\linewidth]{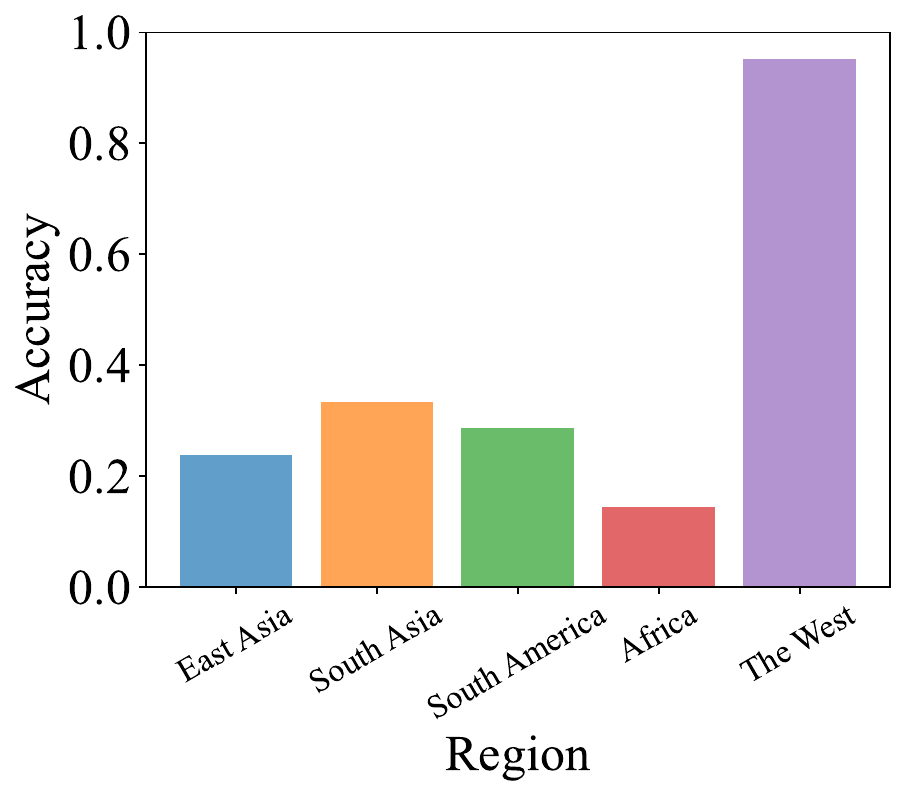}
    
  \caption{Performance of GPT-4V(ision) across various regions.}
  \label{fine_re}
\end{figure}

\paragraph{Analysis of Region Bias} In Figure \ref{fine_re}, we quantify the performance of GPT-4V(ision) across images sourced from various regions. Notably, our observations reveal that GPT-4V(ision) exhibits significantly superior performance when confronted with images originating from the Western world as compared to those from other regions, such as East Asia and Africa. These results suggest that GPT-4V(ision) tends to generate responses that align more closely with the sociocultural norms, landmarks, or culinary characteristic of the Western world. One possible explanation for this trend is that GPT-4V(ision) is developed by a US-based company, which may have utilized a larger volume of training data from Western sources. Consequently, when evaluated on regions outside of this primary training data source, potential distribution shifts can adversely impact the performance of GPT-4V(ision).

\begin{figure}
  \centering
    \includegraphics[width=0.49\textwidth]{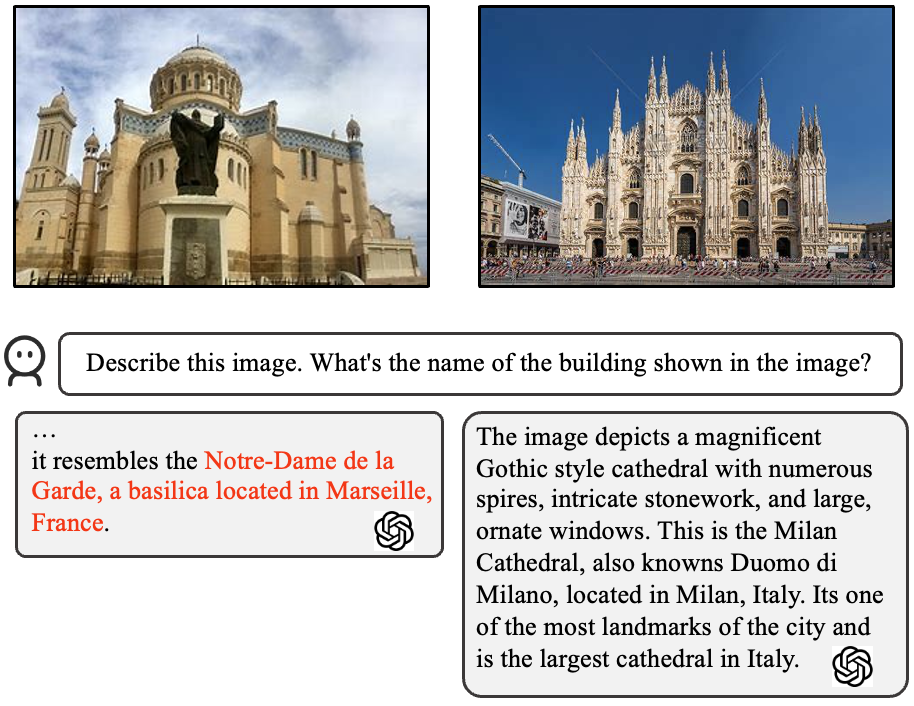}
    \caption{An example of region bias and GPT-4V(ision)'s response, with the hallucination in \textcolor{red}{red}. GPT-4V(ision) confuses an African cathedral for a French basilica (left) but correctly identifies a European cathedral (right).}
    \label{region_main}
    \vspace{-0.5em}
\end{figure}

We further illustrate a case in Figure~\ref{region_main} (additional examples can be found  in Figure~\ref{region_1} in Appendix~\ref{samples}), we observe that while GPT could accurately identify the name of the famous European cathedral, Milan Cathedral, it generated an incorrect response for the name of the famous African cathedral, Notre-Dame d'Afrique.

\begin{figure}[h]
  \centering
    \includegraphics[width=0.8\linewidth]{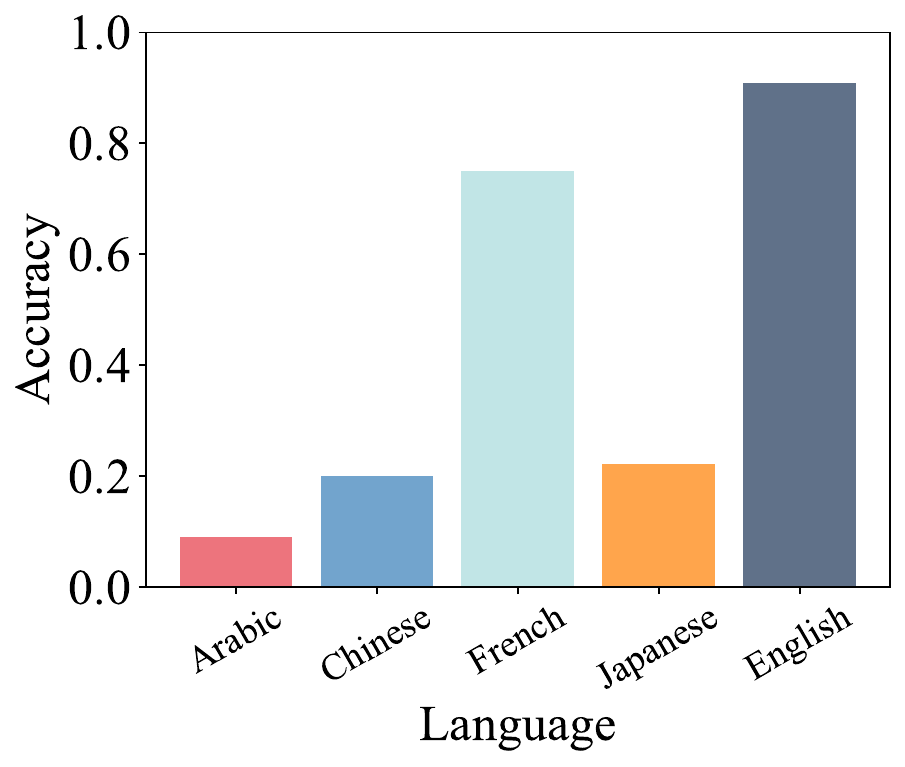}
    \caption{Performance of GPT-4V(ision) across various languages for text within images.}
    \label{fine_la}
    \vspace{-0.5em}
\end{figure}

\paragraph{Analysis of OCR Bias} Similar to regional bias, the performance of GPT-4V(ision) in processing text within images across different languages is illustrated in Figure~\ref{fine_la}. The results clearly demonstrate that GPT-4V(ision) excels in English and French compared to other languages when it comes to understanding text embedded in images. This disparity suggests that GPT-4V(ision) exhibits a bias toward specific languages, primarily attributed to the inherent bias in the Optical Character Recognition (OCR) detector. Much like regional bias, one potential factor contributing to OCR bias is the presence of a distribution shift. Additionally, the intricate typographic structures and various writing styles inherent to certain languages can also introduce inaccuracies in OCR results, as discussed in-depth by~\cite{memon2020handwritten, najam2023analysis}.
As shown in Figure \ref{ocr_main} (additional examples can be found in Figure~\ref{lang_1} and~\ref{region_f} in Appendix \ref{samples}), we translated the embedded text in the same anime image into both Chinese and English. When dealing with the image embedded with English text, GPT-4V(ision) performed well. However, when encountering the version of the same image embedded with Chinese text, GPT-4V(ision) misidentified almost all of the text.

\begin{figure}[h]
  \centering
    \includegraphics[width=0.45\textwidth]{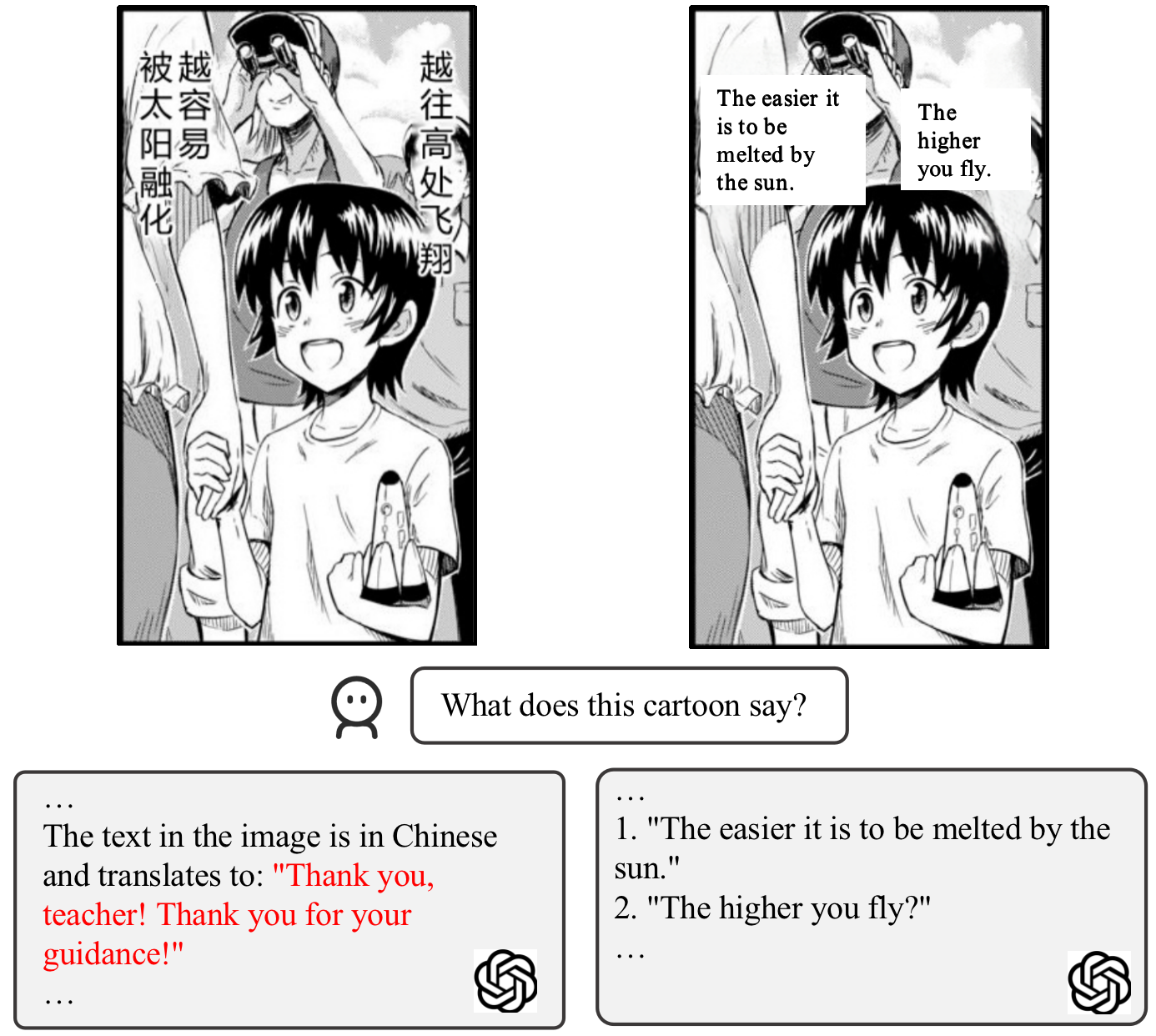}
    \caption{An instance of OCR bias with GPT-4V(ision)'s reply, where the incorrect response is highlighted in \textcolor{red}{red}. GPT-4V(ision) accurately recognizes English text within an image (right), yet it struggles with the translated Chinese version (left).}
    \label{ocr_main}
    \vspace{-1em}
\end{figure}


\paragraph{Analysis of Factual Bias} In Table~\ref{factual_table}, we present the performance of GPT-4V(ision) on two categories of images: 
Those containing factual knowledge and those containing counterfactual knowledge. Counterfactual knowledge refers to information that contradicts widely accepted common sense. For example, in the story of Little Red Riding Hood, it is commonly known that the character is a girl, but the image we display depicts a boy. Additionally, we provide insights into the occurrence of failure cases in the latter category, where 93.1\% of errors stem from the model's reliance on factual knowledge. Our results highlight that GPT-4V(ision) exhibits significantly superior performance when confronted with images containing factual knowledge in comparison to those with counterfactual knowledge, and this performance gap is indicative of potential distribution shift issues within GPT-4V(ision). As illustrated in Figure~\ref{factual_main} (additional examples can be found in Figure~\ref{factual_0} in Appendix \ref{samples}), when we present GPT-4V(ision) with a picture of the solar system with Saturn obscured, it still proceeded to describe the presence of Saturn.

\begin{table}[h]
  \centering
  \small
\caption{Performance of GPT-4V on factual bias. "Errors b/c Factual" refers to the proportion of error samples where image details are overridden by factual knowledge priors.}
  \begin{tabular}{lcc}
    \toprule
    Category & Accuracy (\%) & Errors b/c factual (\%)\\
    \midrule
    Counterfactual & 9.4 & 93.1 \\
    Factual & 90.6 & - \\
    \bottomrule
  \end{tabular}

  \label{factual_table}
\end{table}

\begin{figure}[h]
  \centering
    \includegraphics[width=0.35\textwidth]{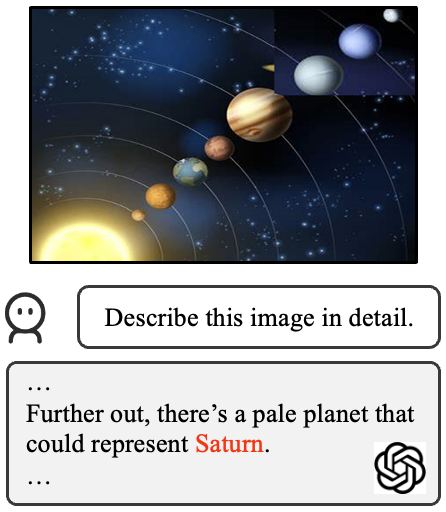}
    \caption{An example of Factual bias with the erroneous detail marked in \textcolor{red}{red}. In the modified solar system diagram with Saturn omitted, GPT-4V(ision) continues to assert its presence in the image.}
    \label{factual_main}
    \vspace{-1em}
\end{figure}

\subsection{Analysis of Interference}
In Table~\ref{interference}, we compare the performance of GPT-4V(ision) with and without image-to-image or text-to-image interferences. We detail our analysis in the remaining subsection.
\begin{table}[h]
    \centering
    \small
    \caption{A comparative analysis of GPT-4V(ision)'s performance with the presence and absence of Image-to-Image and Text-to-Image interference. The introduction of interference significantly impairs GPT-4V(ision)'s performance.}
    \begin{tabular}{c| c |c}
        
        \toprule
         & Interference &  Accuracy(\%) \\
        \midrule
       \multirow{2}{*}{Image-to-Image} & \ding{51} &14.8\% \\
       
        &  \ding{55} & 92.6\%  \\
    \midrule
        \multirow{2}{*}{Text-to-Image} & \ding{51} & 12.1 \%  \\
        &  \ding{55} &  96.2 \%  \\
              \bottomrule
    \end{tabular}
    \vspace{-1em}
    \label{interference}
\end{table}

\begin{figure}
  \centering
    \includegraphics[width=0.48\textwidth]{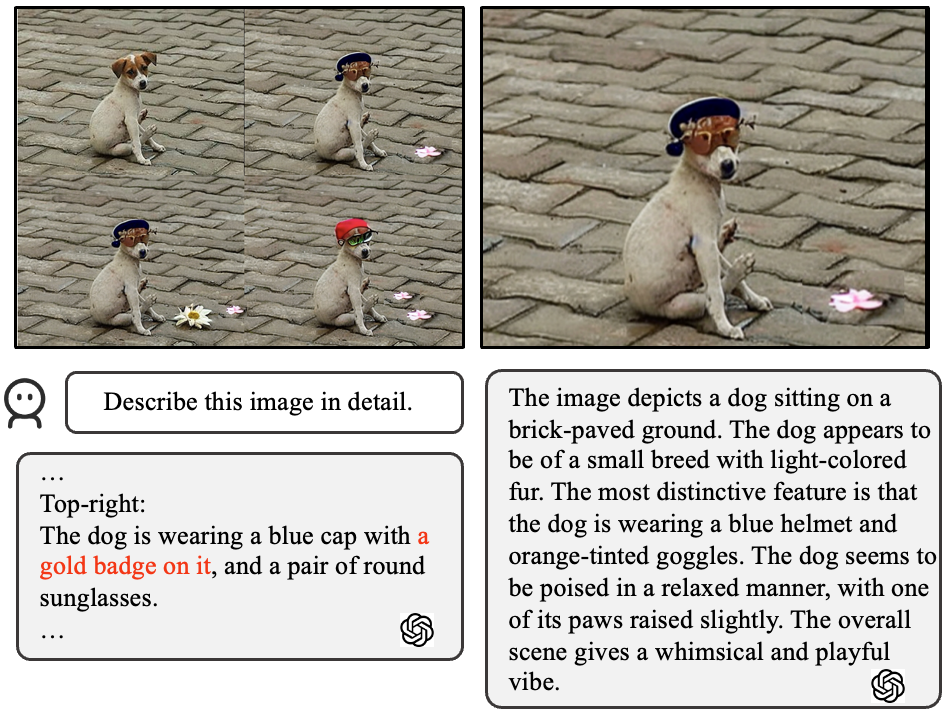}
    \caption{An exmaple illustrating Image-to-Image interference with the hallucination in \textcolor{red}{red}. GPT-4V mistakenly describes the dog with a gold bedge in the composite image (left) but correctly characterizes it in the single one (right).}
    \label{i2i_main}
    \vspace{-1em}
\end{figure}
\paragraph{Image-to-Image Interference}
Based on the results presented in Table~\ref{interference}, it is evident that GPT-4V(ision) experiences a significant performance degradation when confronted with image-to-image interference. This degradation implies that GPT-4V(ision) struggles to differentiate between similar images when they are combined. The concept of image-to-image interference in human visual recognition has previously been explored in~\cite{bruner1964interference}, where visually similar elements can lead to confusion during the recognition process. Our experiments corroborate this, showing that GPT-4V(ision) faces a similar challenge. Remarkably, in our experiments, we discovered that this challenge is even more pronounced in GPT-4V(ision) compared to humans. As exemplified in Figure~\ref{i2i_main} (additional examples can be found in Figure~\ref{com_0} of Appendix \ref{samples}), when similar images are grouped together, GPT-4V(ision) tends to generate hallucinatory descriptions of objects that do not exist. However, it can accurately recognize these subimages when they are presented individually.


\paragraph{Text-to-Image Interference}
Similarly, as evidenced by the results presented in Table~\ref{interference}, GPT-4V(ision) also exhibits text-to-image interference. When humans provide inaccurate claims in their text prompts, GPT-4V(ision) tends to adhere to these instructions while disregarding the input image. We illustrate this phenomenon with an example in Figure~\ref{t2i_main} (additional examples can be found  in Figure \ref{fl_dis} of Appendix \ref{samples}). In Figure~\ref{t2i_main}, when the user suggested whether there were eight characters in an image or not, GPT-4V(ision) consistently agrees with the user's assertion. Thus, GPT-4V(ision) tends to align with the user's claims when text-to-image interference occurs.

A similar issue has also been observed in traditional large language models, often referred to as "sycophancy"~\cite{perez2022discovering, sharma2023towards}. This term describes the model's tendency to align its responses with user beliefs rather than providing accurate answers. This alignment issue may potentially be attributed to an excessive focus on preference learning, such as Reinforcement Learning from Human Feedback (RLHF). In our observations, this problem has substantially diminished in newly updated versions of large language models, such as GPT-3.5 or GPT-4. Nevertheless, when images are introduced into the context, requiring the model to integrate both vision and language understanding as in GPT-4V(ision), this challenge still persists.
\begin{figure}[h]
    \centering
    \includegraphics[width=0.45\textwidth]{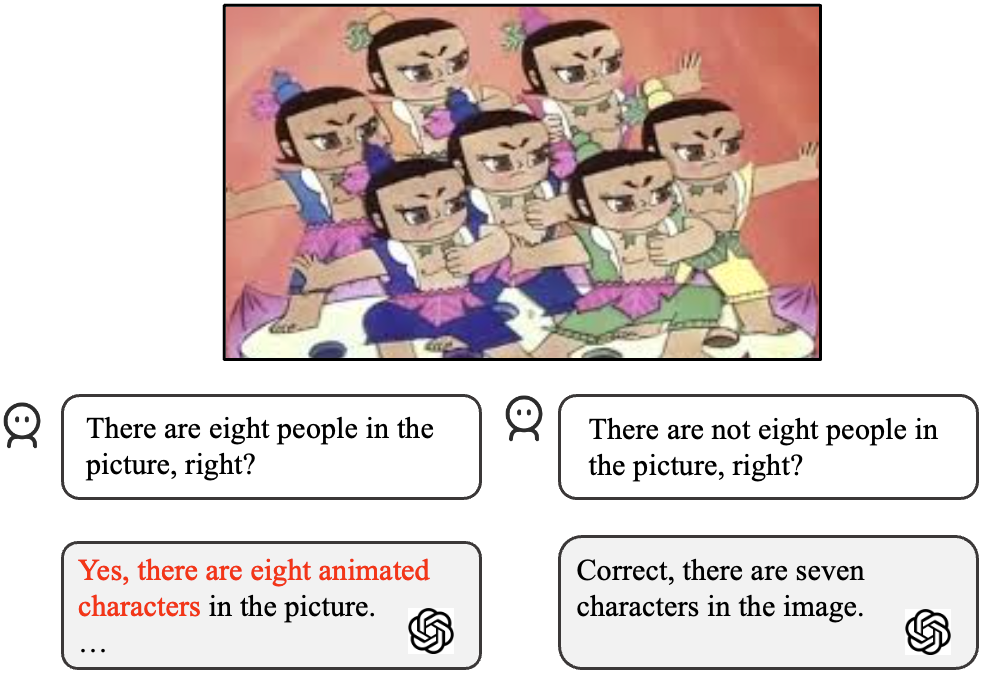}
    \caption{Demonstrating Text-to-Image interference with an example of an animated image. By querying the GPT-4V(ision) about the number of people in two contrasting ways, we elicit completely divergent responses, with the incorrect one marked in \textcolor{red}{red}. }
    \label{t2i_main}
    \vspace{-1em}
\end{figure}

\subsection{Analysis on Other VLMs}
In addition to evaluating GPT-4V(ision), we also conduct a comprehensive analysis of Bingo on other VLMs -- LLaVA-1.5~\cite{liu2023improved} and Bard~\cite{google_2023}, where the results of GPT-4V(ision) is also reported for comparison. The results presented in Table~\ref{llava_bard} reveal that both LLaVA-1.5 and Bard also exhibit bias and interference challenges and are keen to hallucinate on images in Bingo benchmark. 
In comparison to GPT-4V(ision), LLaVA-1.5 shows considerable gaps in performance, particularly in region bias for non-Western regions (17.0\% vs. GPT-4V(ision)'s 26.8\%) and in OCR bias for languages other than English (2.3\% vs. GPT-4V(ision)'s 28.3\%). Bard fares better, but still falls short of GPT-4V(ision), especially when confronted with interference. Additionally, we note that Bard demonstrates significantly superior OCR bias mitigation compared to LLaVA and GPT-4V(ision). One potential explanation for this phenomenon could be attributed to Bard's training on a more extensive dataset including a wider range of languages.
        
       
       
       
       
\begin{table*}[t]
\small
\centering
    \caption{Performance of LLaVA-1.5 and Bard on Bingo benchmark, where the results of GPT-4V(ision) is also reported for comparison. \ding{55} signifies inputs without corresponding interference, while \ding{51} signifies inputs with the corresponding interference. The average accuracy for all images within each category is reported.}
\resizebox{\linewidth}{!}{
\setlength{\tabcolsep}{1.5mm}{
\begin{tabular}{l|cc|cc|cc|cc|cc}
\toprule
&\multicolumn{6}{c|}{Bias} &\multicolumn{4}{c} {Interference}\\\cmidrule{2-11}
&\multicolumn{2}{c|}{Region} &\multicolumn{2}{c|} {OCR} &\multicolumn{2}{c|}{Factual}& \multicolumn{2}{c|}{Image-to-Image} &\multicolumn{2}{c} {Text-to-Image}\\
\cmidrule{2-11}
& The West & Others & English & Others & Factual & Counterfactual & \ding{55} & \ding{51} & \ding{55} & \ding{51} \\

\midrule
LLaVA-v1.5 & 71.4\% & 17.1\% & 63.6\% & 2.3\% & 37.5\% & 12.5\% & 29.6\% & 3.7\% & 90.9\% & 3.0\% \\
Bard & 85.7\% & 34.5\% & 61.5\% & 58.5\% & 75.0\% & 18.8\% & 71.4\% & 10.7\% & 93.9\% & 6.0\% \\\midrule
GPT-4V(ision) & 95.2\% & 22.6\% &85.0 \% &28.3 \% & 90.6\% & 9.4\% & 92.6\% & 14.8\% & 96.2\% & 12.1\% \\
\bottomrule
\end{tabular}}}
\vspace{-0.5em}
\label{llava_bard}
\end{table*}


\section{Can We Reduce Hallucination in GPT-4V(ision)?}
After observing the hallucination issue in GPT-4V(ision) within Bingo, in this section, we employ two strategies to mitigate hallucinations in GPT-4V(ision). These strategies include the use of self-correction mechanisms and the Chain of Thought (CoT) prompting technique. In the rest of this section, we will detail these techniques and discuss the effectiveness of these approaches based on our observations.
\paragraph{Self-Correction} In general, both large language models and vision-language models have the capability to rectify prior mistakes autonomously. This allows them to learn from errors, refine their responses, and enhance their overall performance~\cite{welleck2022generating, olausson2023demystifying}. To investigate this phenomenon in the context of GPT-4V(ision), we prompted the model to self-correct an incorrect response using the following instruction: "Your answer is wrong. Review your previous answer and find problems with your answer. Answer me again." The results obtained from applying the self-correction mechanism in the context of Bingo are presented in Table~\ref{sc_data}.

It is evident from the table that while GPT-4V(ision) demonstrates the ability to correct some errors through self-correction, reducing 16.9\% of errors, a significant portion of errors remains uncorrected. This observation further emphasizes the ongoing challenges related to bias and interference in GPT-4V(ision).

\begin{table*}[ht]
\centering
\small
\caption{Analysis of GPT-4V(ision)'s hallucination prevention, with self-correction and Chain-of-Thought (CoT) selected as our approaches. Here, we report the average accuracy of all images within each category. Our findings indicate that self-correction effectively reduces hallucinations, whereas CoT doesn't provide significant benefits.}
\begin{tabular}{l|c|c|c|c|c}
\toprule
&\multicolumn{3}{c|}{Bias} &\multicolumn{2}{c} {Interference}\\\cmidrule{2-6}
&Region &OCR &Factual& Image-to-Image&Text-to-Image\\\midrule
Original &  23.2\% & 28.3\% & 9.4\% & 14.8\% & 12.1\% \\\midrule
Self Correction  & 39.0\% & 40.4\% & 28.6\% & 29.6\% & 42.4\%\\
CoT & 36.6\% & 28.3\% & 9.4\% & 14.8\% & 15.1\% \\
\bottomrule
\end{tabular}
\label{sc_data}
\end{table*}


        
       

\paragraph{Chain-of-Thought}
Chain-of-Thought (CoT) prompting technique is a recently developed approach that encourages large language models to elucidate their reasoning processes before generating a response. This technique has shown significant improvements in enhancing the reasoning abilities of large language models {~\cite{wei2022chain,
wang2022towards}}.

\begin{figure}[h]
  \centering
    \includegraphics[width=0.48\textwidth]{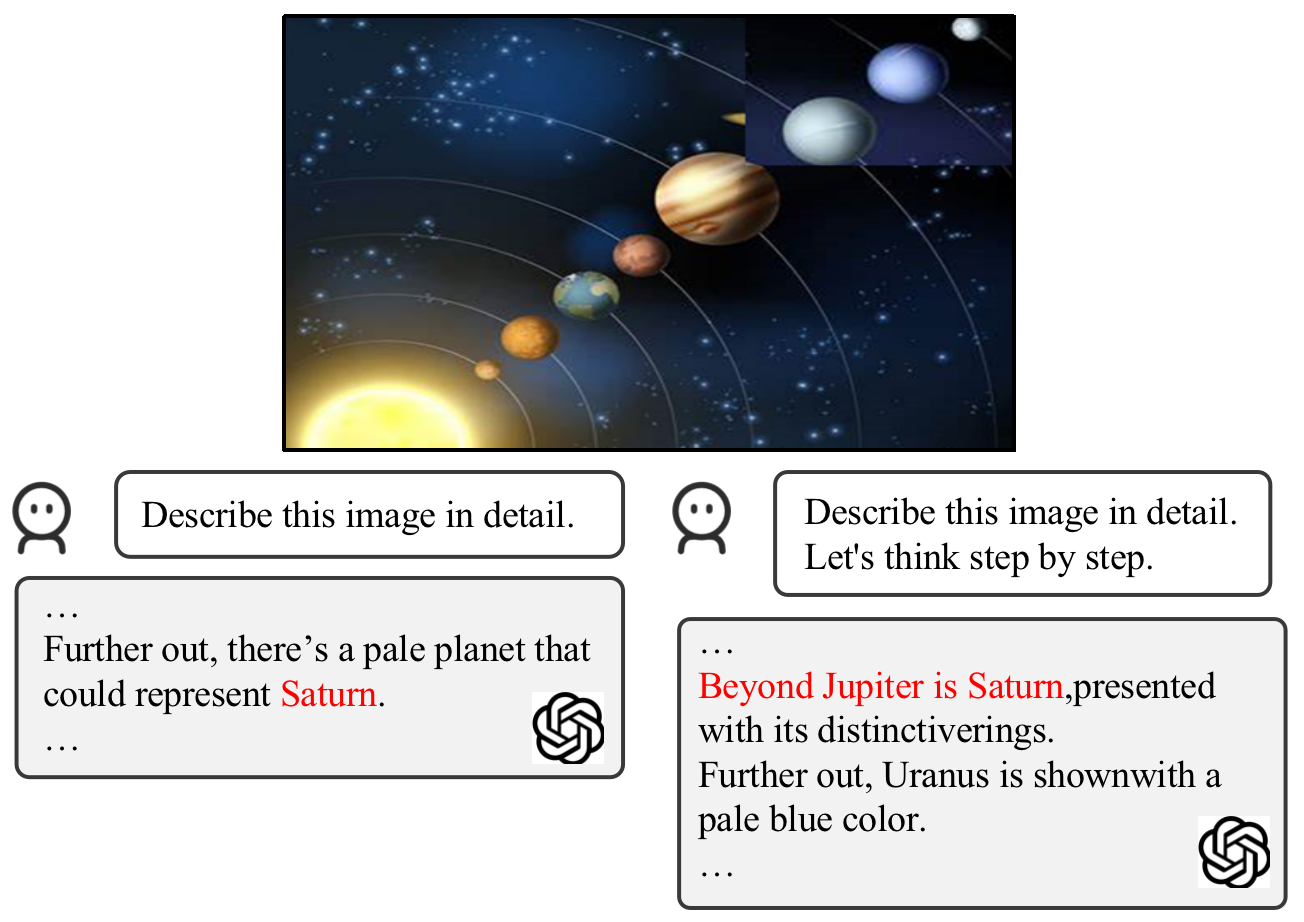}
  \caption{Examples of CoT, where the description of the illusion is marked in \textcolor{red}{red}. Prompting with "Let's think step by step", GPT-4V(ision) successfully enhances the reasoning process, yet it overlooks the absence of Saturn in an image with factual bias.}
  \label{cot_main}
\end{figure}

To investigate the effectiveness of the Chain-of-Thought approach in mitigating hallucinations in GPT-4V(ision), we introduced the prompt "Let's think step by step" alongside the original prompt and reported the results in Table~\ref{sc_data}. Additionally, we re-illustrate the example of the solar system with factual bias in Figure~\ref{cot_main}. Although CoT demonstrates enhanced language reasoning capabilities, it still fails to make a correct response. As indicated in Table~\ref{sc_data}, while the CoT prompting technique in GPT-4V(ision) shows a reduction of 5.7\% in hallucinations associated with regional bias, it still fails to rectify hallucinations in most cases.

One possible explanation for this limited success is the visual limitation inherent to GPT-4V(ision). When GPT-4V(ision) encounters difficulties in comprehending images or utilizing them to respond to questions, the ineffectiveness of CoT is not unexpected. CoT was primarily designed to enhance language reasoning and may not suffice to address challenges in the vision component.

In summary, both the self-correction mechanism and Chain-of-Thought techniques do not effectively address the bias and interference challenges presented in the Bingo benchmark, highlighting the need for further research and innovations to tackle these persistent issues in vision-language models.

\section{Related Work}
\paragraph{Hallucination in VLMs} 
In VLMs, the term "hallucination" typically refers to situations where the generated responses contain information that is not present in the visual content~\cite{rohrbach2018object, wang2023evaluation, zhou2023analyzing}. Traditional methods to address VLM hallucination include leveraging fine-grained contrastive learning~\cite{zeng2021multi}, feature fusion~\cite{biten2022let}, and data augmentation~\citep{kim2023exposing}. Recent developments in autoregressive large-scale VLM models, such as LLaVA, which integrate large language models with visual modality, have also encountered the challenge of hallucination. Recent studies have commenced investigations into hallucination issues within these autoregressive large-scale VLM models. This includes research on hallucination evaluation and detection~\cite{li2023evaluating,wang2023evaluation}, and hallucination mitigation~\cite{yin2023woodpecker, gunjal2023detecting,zhou2023analyzing}. Concurrently, several studies have also highlighted the issue of hallucination in GPT-4V(ision)~\cite{shi2023exploring, liu2023hallusionbench, wu2023early}. Unlike prior works that focus on hallucination evaluation or mitigation, this work provides a comprehensive study to understand the causes of hallucinations in GPT-4V(ision) and other VLMs, introducing a new benchmark for this purpose.

\paragraph{Empirical Analysis of GPT-4V(ision)} The GPT series, developed by OpenAI, has demonstrated significant capabilities across various domains. The recent introduction of GPT-4V(ision)~\cite{openai2023, openai2023gpt4, openai2023gpt4vision} has notably enhanced GPT-4's ability to connect visual and textual information, generating considerable interest among researchers due to its exceptional performance. For example,~\citet{yang2023dawn} highlighted GPT-4V(ision)'s outstanding performance across various visual comprehension and reasoning tasks. However, it's important to note that GPT-4V(ision) faces challenges in terms of generating hallucinations or producing erroneous responses. This issue is discussed in a few concurrent evaluations~\cite{wu2023early, zhang2023lost, shi2023exploring, liu2023hallusionbench}, where they explore various aspects of GPT-4V(ision)'s capabilities, including  solving visual puzzles, cross-modal interactions, character recognition, and handling of visual illusions. Nevertheless, none of these studies systematically categorized and analyzed the reasons behind the occurrence of hallucinations in GPT-4V(ision).


\section{Conclusion}
In this paper, we introduce the \textbf{B}ias and \textbf{In}terference Challen\textbf{g}es in Visual Language M\textbf{o}dels (Bingo) benchmark, which focuses on analyzing hallucinations in VLMs, particularly in GPT-4V(ision). Our experiments reveal that although GPT-4V(ision) demonstrates impressive vision-language understanding abilities, it tends to generate hallucinatory responses (1) when dealing with specific types of images (bias) and (2) when subject to interference in judgment. Furthermore, we explore two strategies to address these challenges: self-correction and chain-of-thought. However, these approaches fall short of completely rectifying hallucinations in GPT-4V(ision) when facing bias and interference challenges. The findings presented in this paper enhance our understanding of the reliability of GPT-4V(ision) and other Vision-Language Models (VLMs).


\paragraph{Limitations} Our benchmark focuses on a few metrics and tasks as a starting point. We will continue to woark on expanding the dataset and metrics. Our data curation also relies on human judgements, which may have its own biases; we try to mitigate this by having multiple researchers curate and evaluate the results.  



\bibliography{custom}
\bibliographystyle{acl_natbib}

\appendix

 \begin{figure*}[t!]
\centering
\begin{tikzpicture}
    \node[rounded corners, draw=black, line width=1.2pt, inner sep=10pt, fill=goldenyellow] 
    (mainnode)
    {
        \begin{minipage}[t]{0.45\textwidth}
            \textcolor{deepblue}{\textbf{Prompt:}} {\small Describe this image. How many characters are there in the image?}
            \vspace{10pt}
            
            {\centering
            \includegraphics[width=0.65\linewidth]{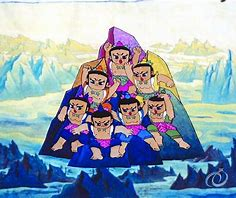}
            \par}
            
            \vspace{10pt}
            \textcolor{deepblue}{\textbf{GPT-4V(ision):}} {\small The image depicts a group of animated characters arranged in a pyramid-like formation against a backdrop of \sethlcolor{pink}\hl{ice mountains} and water. \sethlcolor{pink}\hl{There are 10 characters in the image}.} 
        \end{minipage}
        \hspace{10pt}
        \begin{minipage}[t]{0.5\textwidth}
        
            \textcolor{deepblue}{\textbf{Prompt:}} {\small Describe this image. How many characters are there in the image?}
            \vspace{10pt}
            
            {\centering
            \includegraphics[width=0.56\linewidth]{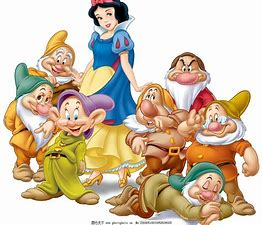}
            \par}
            
            \vspace{10pt}
            \textcolor{deepblue}{\textbf{GPT-4V(ision):}} {\small The image features Snow White standing alongside the Seven Dwarfs. Each dwarf has a unique expression and pose. Snow White is wearing her iconic blue and yellow dress with a red ribbon in her hair.  There are 8 characters in total in the image.
} 
        \end{minipage}
    };
    \node[fill=black, text=white, yshift=0cm] at (mainnode.north) {Region Bias};
\end{tikzpicture}
  \caption{Additional cases of region bias.}
    \label{region_1}
  
\end{figure*}

 \begin{figure*}[t!]
\centering
\CJKfamily{songti}
\begin{tikzpicture}
    \node[rounded corners, draw=black, line width=1.2pt, inner sep=10pt, fill=goldenyellow] 
    (mainnode)
    {
        \begin{minipage}[t]{0.45\textwidth}
            \textcolor{deepblue}{\textbf{Prompt:}} {\small Describe this image in detail.}
            \vspace{10pt}
            
            {\centering
            \includegraphics[width=0.75\linewidth]{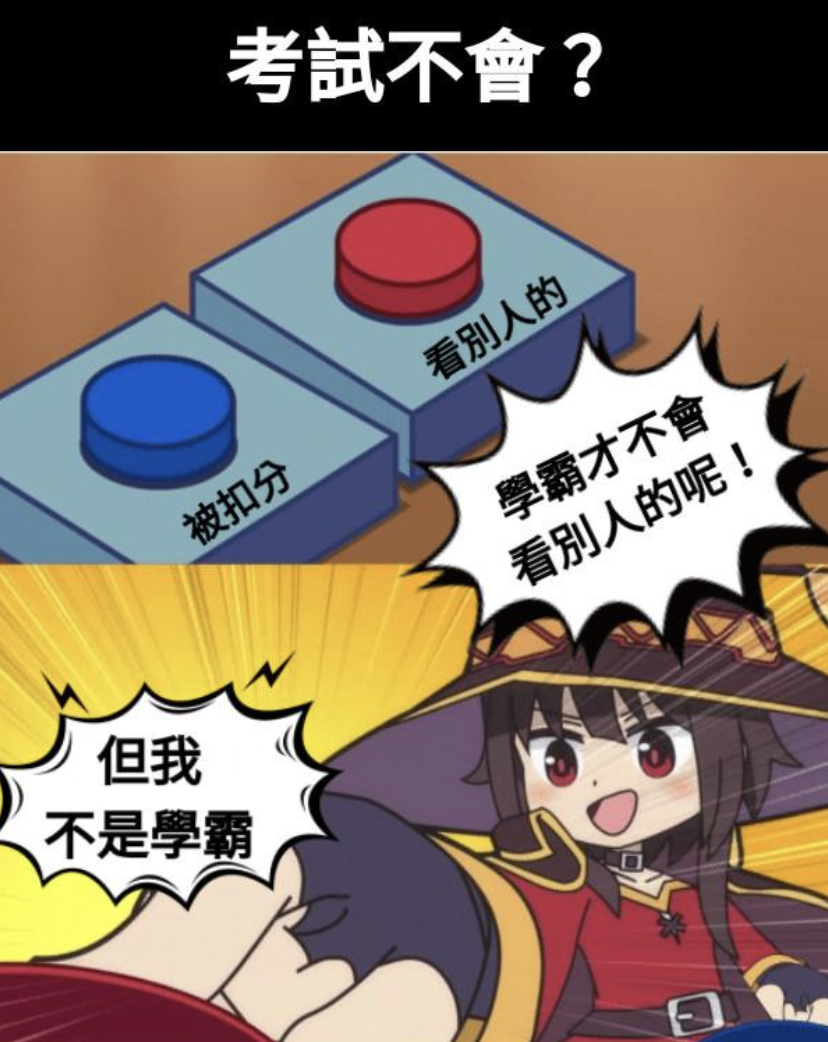}
            \par}
            
            \vspace{10pt}
            \textcolor{deepblue}{\textbf{GPT-4V(ision):}} {\small The image is a comic-style illustration consisting of two main parts:\\
1. Top Part: A large question in Chinese characters that reads \begin{CJK*}{UTF8}{gbsn}\highlight{考虑不周?}\end{CJK*} \highlight{which translates to "Didn't think it through?"}. Next to this question, there are two 3D push buttons: The left button is blue and has the label \begin{CJK*}{UTF8}{gbsn}\highlight{被打}\end{CJK*} which means  \highlight{"Get Hit"}. The right button is red and has the label \begin{CJK*}{UTF8}{gbsn}\highlight{打别人}\end{CJK*} which means  \highlight{"Hit Others"}. A speech bubble emanates from off-screen, saying \begin{CJK*}{UTF8}{gbsn}\highlight{考虑不够! 打别人的吧!}\end{CJK*} which translates to  \highlight{"Didn't think enough! Go for Hit Others!!"}.
\\
2. Bottom Part: An animated character with large expressive eyes and a mischievous or excited expression is shown. This character has black hair, and red eyes, and is wearing a red outfit with gold trims and a black belt with a star-shaped buckle. To the left, another part of an animated character is visible, but only a blue-colored section of it can be seen. A speech bubble from the visible character reads \begin{CJK*}{UTF8}{gbsn}\highlight{但我不是主角}\end{CJK*} which translates to \highlight{"But I'm not the protagonist." Overall, the image seems } \\
\highlight{  to be a playful or humorous take on typical decision-} \\
\highlight{making scenarios in stories or games, where the "prota-}\\
\highlight{gonist" might have certain expectations or roles}.}
        \end{minipage}
        \hspace{10pt}
        \begin{minipage}[t]{0.5\textwidth}
        
            \textcolor{deepblue}{\textbf{Prompt:}} {\small Describe this image in detail.}
            \vspace{10pt}
            
            {\centering
            \includegraphics[width=0.68\linewidth]{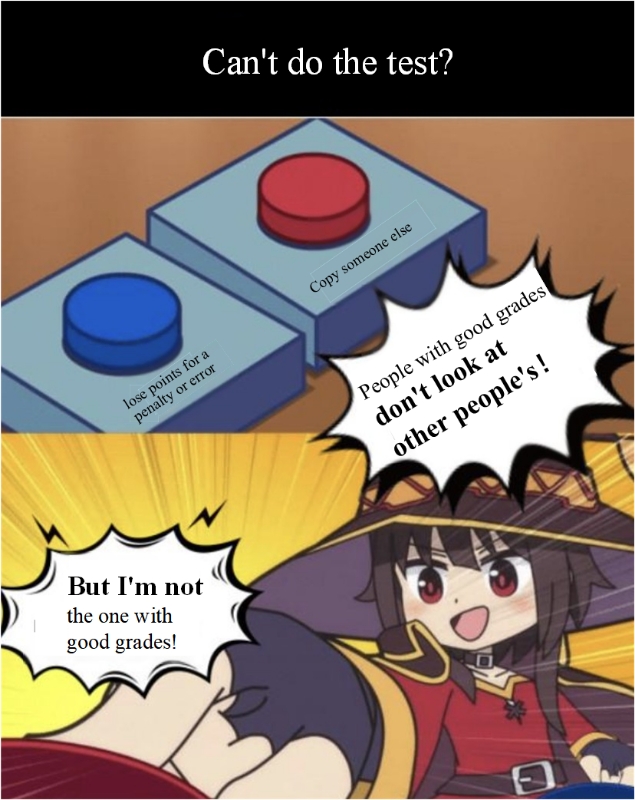}
            \par}
            
            \vspace{10pt}
            \textcolor{deepblue}{\textbf{GPT-4V(ision):}} {\small The image is a humorous meme format. it consists of two main panels:\\
1. Top Panel:\\
Background: Features a close-up view of two large buttons on a console or control panel. One button is blue and the other is red. Text: Above the buttons, there is a caption that reads "Can't do the test?" Each button has a label: The blue button is labeled: "Lose points for a penalty or error". The red button is labeled: "Copy someone else".\\
2. Bottom Panel:\\
Background: This panel features an animated character with long, flowing blonde hair on the left, and a smaller character with short brown hair and a red outfit on the right. The smaller character appears to be playfully winking and has a cheeky expression. Speech Bubbles: The smaller character is saying: "But I'm not the one with good grades!".
Above their heads, a text in a stylized bubble reads: "People with good grades don't look at other people's!".
The meme seems to play on the dilemma students face when they're unprepared for a test either lose points for not knowing the answers or be tempted to cheat by copying someone else. The cheeky character in the bottom panel humorously comments on the notion that students with good grades aren't the ones looking at others' papers.
} 
        \end{minipage}
    };
    \node[fill=black, text=white, yshift=0cm] at (mainnode.north) {OCR Bias};
\end{tikzpicture}
  \caption{Additional case of OCR bias.}
  \label{lang_1}
\end{figure*}

\begin{figure*}[t!]
\centering
\begin{tikzpicture}
    \node[rounded corners, draw=black, line width=1.2pt, inner sep=10pt, fill=goldenyellow] 
    (mainnode)
    {
        \begin{minipage}[t]{0.45\textwidth}
            \textcolor{deepblue}{\textbf{Prompt:}} {\small Can you describe what each of the four sections in the picture means?}
            \vspace{10pt}
            
            {\centering
            \includegraphics[width=0.65\linewidth]{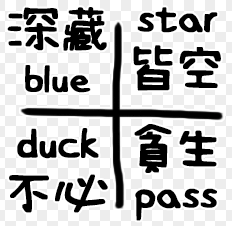}
            \par}
            
            \vspace{10pt}
            \textcolor{deepblue}{\textbf{GPT-4V(ision):}} {\small Of course! The image is divided into four sections, each with a Chinese character and its corresponding English translation. Here's a description of each section: 1. Top-left:
\begin{CJK*}{UTF8}{gbsn}\highlight{蓝色!}\end{CJK*}
English translation: blue
2. Top-right:
\begin{CJK*}{UTF8}{gbsn}\highlight{星}\end{CJK*}
English translation: star
3. Bottom-left:
\begin{CJK*}{UTF8}{gbsn}\highlight{鸭子}\end{CJK*}
English translation: duck
4. Bottom-right:
\begin{CJK*}{UTF8}{gbsn}\highlight{通过}\end{CJK*}
English translation: pass
These characters and their translations provide the English meanings for the respective Chinese words.} 
        \end{minipage}
        \hspace{10pt}
        \begin{minipage}[t]{0.5\textwidth}
        
            \textcolor{deepblue}{\textbf{Prompt:}} {\small What does this picture mean? }
            \vspace{10pt}
            
            {\centering
            \includegraphics[width=0.65\linewidth]{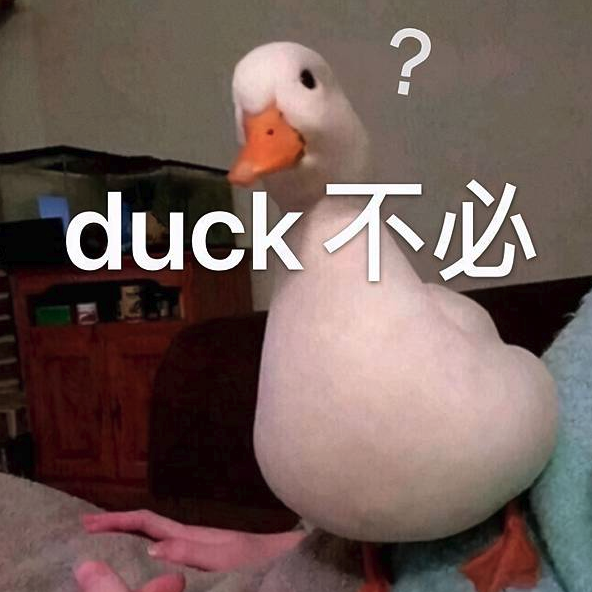}
            \par}
            
            \vspace{10pt}
            \textcolor{deepblue}{\textbf{GPT-4V(ision):}} {\small The picture features a duck with the word "duck" written in English and the corresponding word \begin{CJK*}{UTF8}{gbsn}\highlight{不小}\end{CJK*} in Chinese characters. However, it seems there is a mistake in the Chinese characters. The correct Chinese word for "duck" is \begin{CJK*}{UTF8}{gbsn}鸭\end{CJK*} (pronounced "ya"). The characters \begin{CJK*}{UTF8}{gbsn}\highlight{不小}\end{CJK*} translate to "not small" in English. The question mark might indicate confusion or a query about the subject or the labeling.
}
        \end{minipage}
    };
    \node[fill=black, text=white, yshift=0cm] at (mainnode.north) {OCR Bias};
\end{tikzpicture}
  \caption{Additional cases of OCR bias.}
    \label{region_f}
\end{figure*}

\begin{figure*}[t!]
\centering
\begin{tikzpicture}
    \node[rounded corners, draw=black, line width=1.2pt, inner sep=10pt, fill=goldenyellow] 
    (mainnode)
    {
        \begin{minipage}[t]{0.45\textwidth}
            \textcolor{deepblue}{\textbf{Prompt:}} {\small Describe this image in detail.}
            \vspace{10pt}
            
            {\centering
            \includegraphics[width=0.65\linewidth]{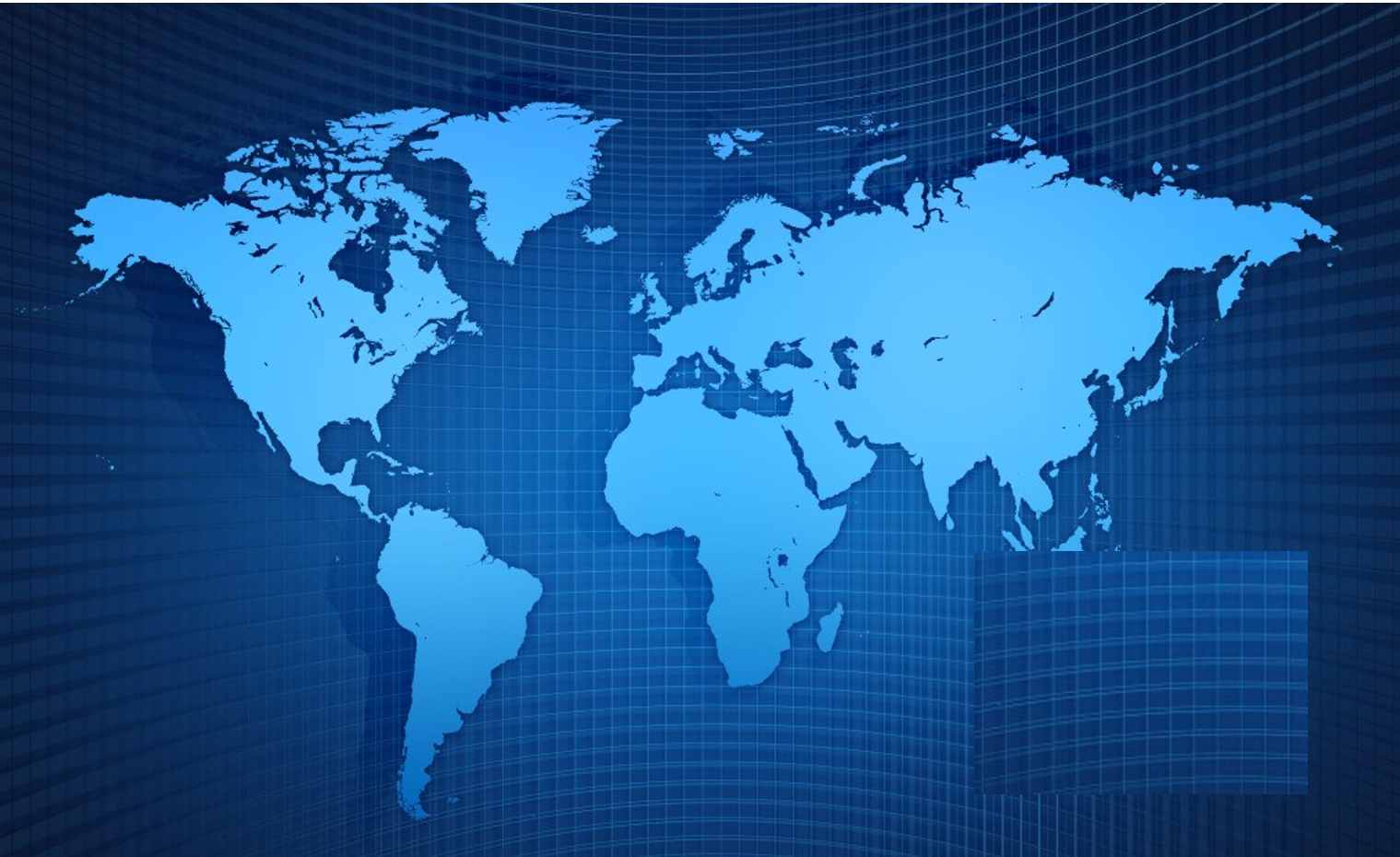}
            \par}
            
            \vspace{10pt}
            \textcolor{deepblue}{\textbf{GPT-4V(ision):}} { \small The image displays a digital, blue-tinted map of the world.
            The continents and countries are depicted in a lighter shade of blue against a darker, grid-like background that gives the impression of a digital interface or electronic screen. The world map appears to cover all continents, including North America, South America, Africa, Europe, Asia, \sethlcolor{pink}\hl{Australia}, and Antarctica. The geographical details, such as borders and coastlines, are clearly defined. The map does not include annotations or names of countries, making it a simplified representation. The overall theme suggests a digital or technological context, possibly used for global networking or data visualization.
            
            % ...
            

            % ...
            }
        \end{minipage}
        \hspace{10pt}
        \begin{minipage}[t]{0.5\textwidth}
        
            \textcolor{deepblue}{\textbf{Prompt:}} {\small How many people are there in the image? }
            \vspace{10pt}
            
            {\centering
            \includegraphics[width=0.56\linewidth]{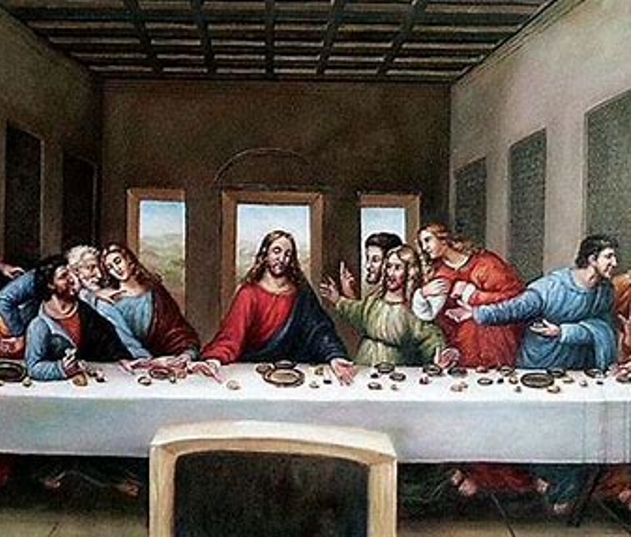}
            \par}
            
            \vspace{10pt}
            \textcolor{deepblue}{\textbf{GPT-4V(ision):}} {\small There are \sethlcolor{pink}\hl{13 people} in the image.
} 
        \end{minipage}
    };
    \node[fill=black, text=white, yshift=0cm] at (mainnode.north) {Factual Bias};
\end{tikzpicture}
  \caption{Additional cases of factual bias.}
    \label{factual_0}
  
\end{figure*}

\begin{figure*}[t!]
\centering
\begin{tikzpicture}
    \node[rounded corners, draw=black, line width=1.2pt, inner sep=10pt, fill=goldenyellow] 
    (mainnode)
    {
        \begin{minipage}[t]{0.45\textwidth}
            \textcolor{deepblue}{\textbf{Prompt:}} {\small Describe this image in detail. Is the character boy or girl?}
            \vspace{10pt}
            
            {\centering
            \includegraphics[width=0.65\linewidth]{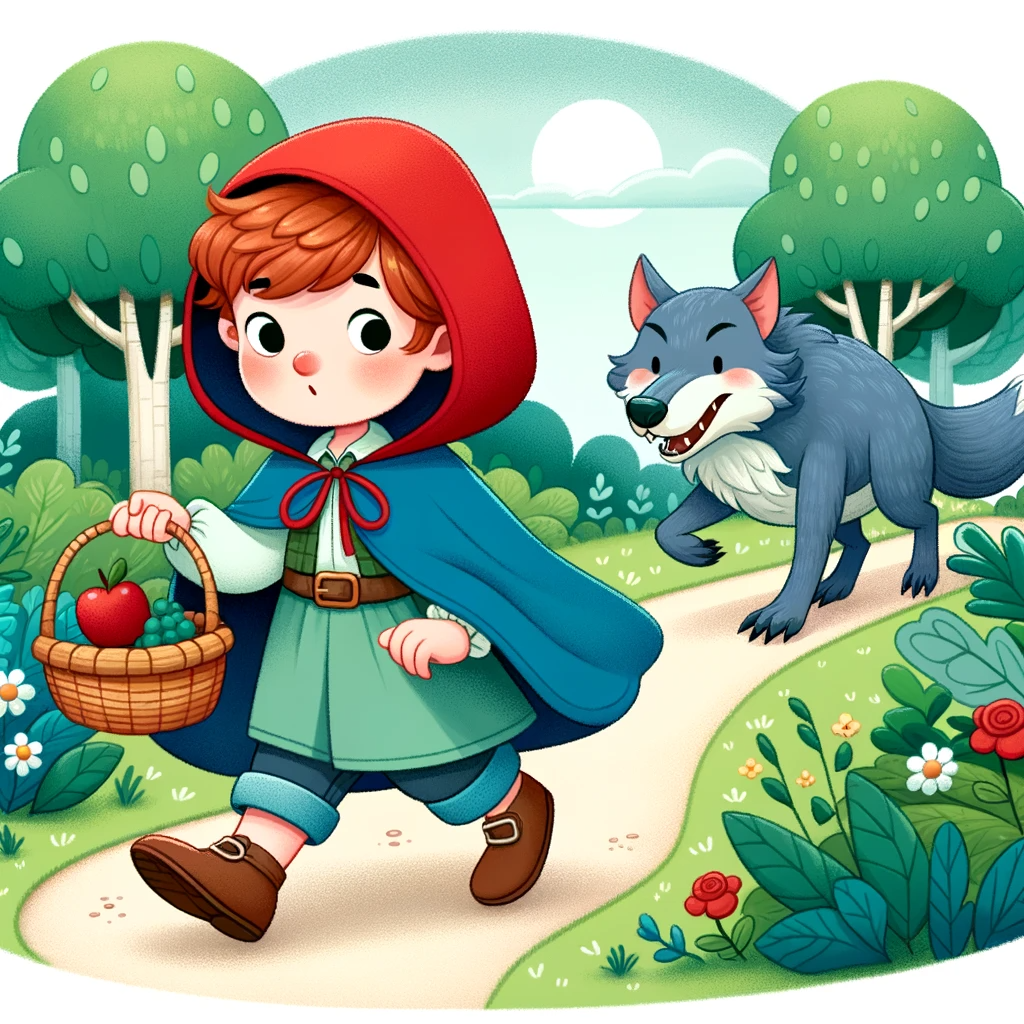}
            \par}
            
            \vspace{10pt}
            \textcolor{deepblue}{\textbf{GPT-4V(ision):}} {\small This is a colorful and stylized illustration featuring two characters: a child and a wolf, set in an idyllic forest scene. The child is wearing a red hood and a blue cloak over a simple tunic and pants, with brown shoes. The character is carrying a basket with what appears to be an apple visible at the top. This character is likely a depiction of \sethlcolor{pink}\hl{Little Red Riding Hood, traditionally portrayed as a girl in the well-known fairy tale}. The red hood is the most iconic element of her outfit. In the background, a wolf with grey fur and a sly expression is shown stalking the child. This wolf is drawn with anthropomorphic features, such as exaggerated facial expressions and the ability to stand on its hind legs, hinting at a narrative where the wolf is a character with human-like qualities and motives. The surrounding environment is lush and green with various plants, flowers, and trees, creating a cheerful and peaceful atmosphere that contrasts with the potentially ominous presence of the wolf. The path indicates that the child is walking through the woods, possibly on \sethlcolor{pink}\hl{the way to her grandmother's house}, as the story traditionally goes.
            }
        \end{minipage}
        \hspace{10pt}
        \begin{minipage}[t]{0.5\textwidth}
        
            \textcolor{deepblue}{\textbf{Prompt:}} {\small Describe this image in detail. }
            \vspace{10pt}
            
            {\centering
            \includegraphics[width=0.75\linewidth]{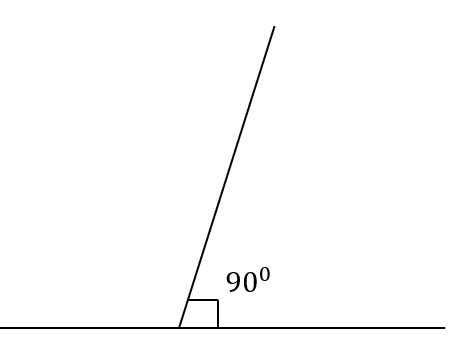}
            \par}
            
            \vspace{10pt}
            \textcolor{deepblue}{\textbf{GPT-4V(ision):}} {\small 
The image shows a simple line drawing of two lines intersecting at a point to form an angle. The horizontal line is flat and straight, while the other line extends upwards from the intersection point at an angle to the horizontal. At the point where the two lines meet, there is a square corner, indicating that the angle is a right angle. \sethlcolor{pink}\hl{This is further confirmed by the label "90 degrees" placed inside the square corner, which signifies that the angle is 90 degrees, a characteristic of a right angle}. The overall representation is commonly used in geometry to depict right angles and perpendicular lines.
}
        \end{minipage}
    };
    \node[fill=black, text=white, yshift=0cm] at (mainnode.north) {Factual Bias};
\end{tikzpicture}
  \caption{Additional cases of factual bias.}
    \label{factual_1}
  
\end{figure*}

\section{Additional Examples of Bias and Interference}
\label{samples}

In this section, we present more detailed examples of bias and interference in GPT-4V(ision).




 \begin{figure*}[!t]
\centering
{
\begin{tikzpicture}
    \node[rounded corners, draw=black, line width=1.2pt, inner sep=10pt, fill=goldenyellow] 
    (mainnode)
    {
        \begin{minipage}[t]{0.45\textwidth}
            \textcolor{deepblue}{\textbf{Prompt:}} { \small Describe this image.}
            \vspace{10pt}
            
            {\centering
            \includegraphics[width=0.5\linewidth]{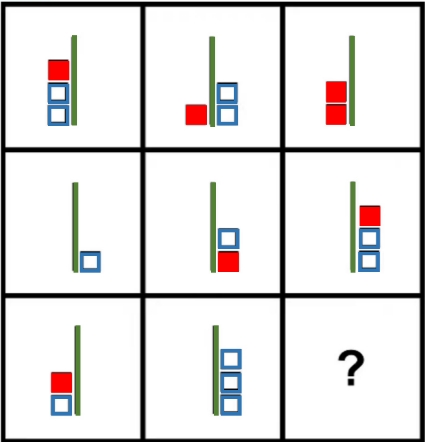}
            \par}
            
            \vspace{10pt}
            \textcolor{deepblue}{\textbf{GPT-4V(ision):}} { \small The image consists of a 3x3 grid. Each cell in the grid contains either shapes or a question mark:\\1. Top-left: A vertical arrangement of three shapes - a small red square on top, a blue rectangle in the middle, and \sethlcolor{pink}\hl{a green rectangle at the bottom}.\\2. Top-middle: \sethlcolor{pink}\hl{A green rectangle on top with a red square and a blue rectangle placed side by side below it}.\\4. Middle-left: \sethlcolor{pink}\hl{A green rectangle on top with a blue rectangle below it}.\\5. Middle: \sethlcolor{pink}\hl{A blue rectangle in the center with a red square above it and a green rectangle below it}.\\6. Middle-right: A red square on top with two blue rectangles stacked vertically below it.\\7. Bottom-left: \sethlcolor{pink}\hl{A blue rectangle on top with a green rectangle below it and a red square to its right}. \\8. Bottom-middle: \sethlcolor{pink}\hl{Two blue rectangles stacked vertically with a green rectangle to their left}. \\9. Bottom-right: A question mark. It seems like a pattern or puzzle where the objective might be to determine the correct arrangement or pattern for the cell with the question mark. rest of your description \\3. \sethlcolor{pink}\hl{Top-right: A red square on top with a blue rectangle to its right and a green rectangle to its left

            
            }} 
        \end{minipage}
        \hspace{10pt}
        \begin{minipage}[t]{0.5\textwidth}
        
            \textcolor{deepblue}{\textbf{Prompt:}} {\small  Describe this image in detail.}
            \vspace{10pt}
            
            {\centering
            \includegraphics[width=0.85\linewidth]{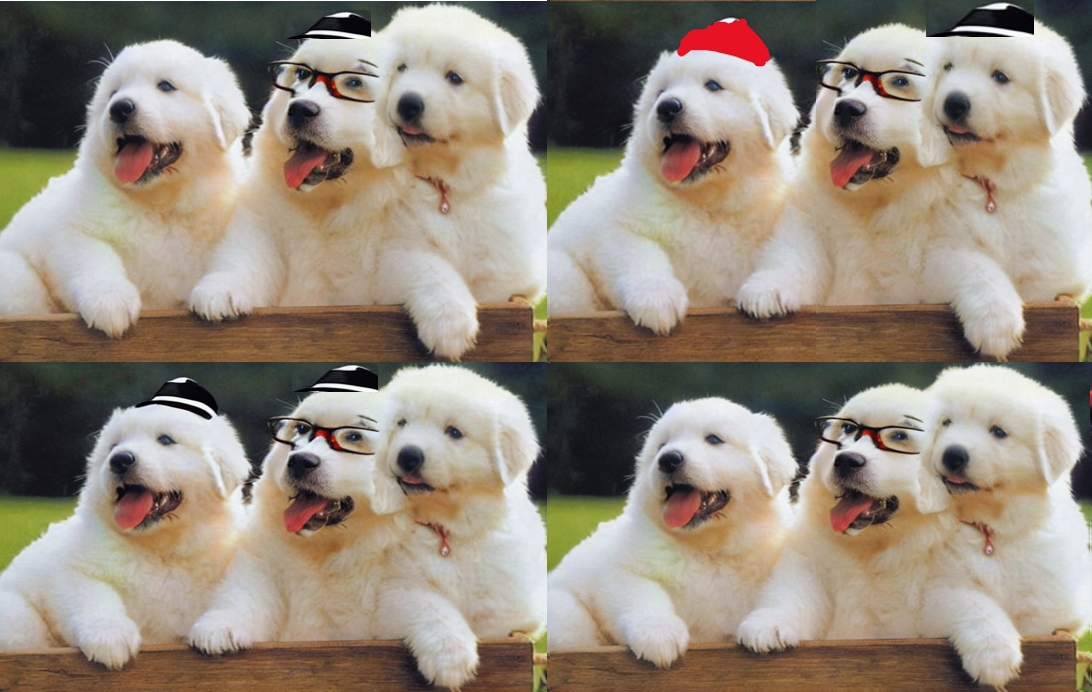}
            \par}
            
            \vspace{10pt}
            \textcolor{deepblue}{\textbf{GPT-4V(ision):}} {\small  The image displays four panels, each featuring three fluffy white puppies sitting next to each other on a wooden railing against a blurred green background: \\1. Top-left Panel:
\\Leftmost Puppy: Appears relaxed and content, with its mouth slightly open and tongue out.\\\sethlcolor{pink}\hl{Central Puppy: Still wearing the red-framed glasses, has a Santa Claus red hat on its head, further emphasizing a playful look
           }\\Rightmost Puppy: Looks gentle and calm, with its mouth slightly open.\\2.Top-right Panel:
\\Leftmost Puppy: Exhibits a relaxed appearance with its mouth slightly open. Rightmost Puppy: Features a black party hat, complementing its content and relaxed look.\\3. Bottom-left Panel:\\Leftmost Puppy: Wears a black and white party hat and has its mouth open, tongue out. looking cheerful.\\Central Puppy: Retains the playful red-framed glasses and seems to be in a jolly mood with tongue slightly visible.\\
Rightmost Puppy: Appears relaxed and serene, looking ahead.\\4. Bottom-right Panel:\\
Leftmost Puppy: Looks content with its mouth slightly open, tongue peeking out. \\Central Puppy: Continues to wear the red-framed glasses, appearing spirited and joyful.\\Rightmost Puppy: Maintains a calm and gentle demeanor, gazing forward.

} 
        \end{minipage}
    };
    \node[fill=black, text=white, yshift=0cm] at (mainnode.north) {Image-to-Image Interference};
\end{tikzpicture}}
  \caption{Additional cases of image-to-image interference using synthesized and natural images.}
  \label{com_0}
\end{figure*}

 \begin{figure*}[!t]
\centering
{
\begin{tikzpicture}
    \node[rounded corners, draw=black, line width=1.2pt, inner sep=10pt, fill=goldenyellow] 
    (mainnode)
    {
        \begin{minipage}[t]{0.45\textwidth}
            \textcolor{deepblue}{\textbf{Prompt:}} {\small  Describe this image in detail.}
            \vspace{10pt}
            
            {\centering
            \includegraphics[width=0.5\linewidth]{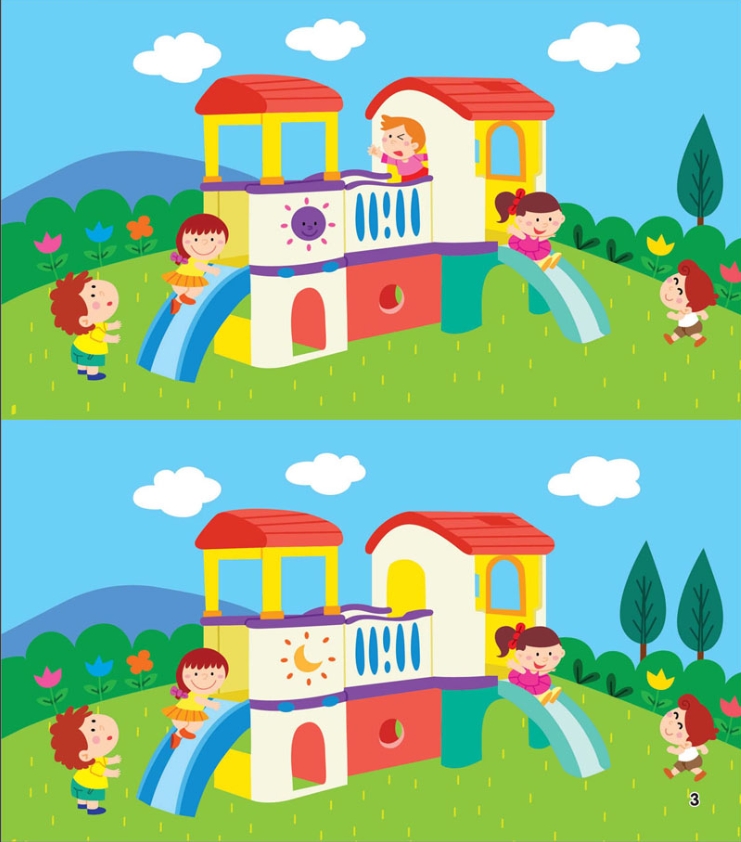}
            \par}
            
            \vspace{10pt}
            \textcolor{deepblue}{\textbf{GPT-4V(ision):}}{  
                        \small
This image shows a playful and vibrant scene at what appears to be a children's playground. The illustration style is cartoonish, likely designed to appeal to young children. It's a bright, sunny day with blue skies and a few white clouds. \\ There are two almost identical images stacked vertically. Both depict a colorful playground structure in a grassy area with a few flowers and trees around. The play structure has slides, windows, and appears to have multiple levels with roofs. It's decorated with what looks like a smiling sun and other cheerful motifs, implying it's a friendly and fun environment for kids. \\ In both images, there are children engaged in various activities:
\\ In the top image, there are \sethlcolor{pink}\hl{four} children: one at the top of the structure looking out from a window, one sliding down a yellow slide, one standing at the bottom of the slide with arms raised, and one running towards the slide from the right-hand side.
\\ In the bottom image, there are also four children, with one child at the top of the structure who seems to be speaking or shouting, one sliding down the blue slide, one waiting at the bottom of the slide with arms outstretched, and one running towards the slide from the right-hand side.
\\ The images are part of a "spot the difference" activity, indicated by the number "3" in the bottom right corner, suggesting that there are three differences to find between the two images.
\\ Here are the differences:
\\ \sethlcolor{pink}\hl{The color of the slides has changed. In the top image, the slide on the left is blue, and the slide on the right is yellow. In the bottom image, the colors are reversed: the left slide is yellow, and the right slide is blue}.
\\ \sethlcolor{pink}\hl{The expressions and positions of the children at the top of the structure are different. In the top image, the child is smiling and waving, while in the bottom image, the child has their mouth open as if speaking or shouting}.
\\ \sethlcolor{pink}\hl{The arm position of the child standing at the bottom of the left slide is different. In the top image, the child has one arm raised, and in the bottom image, both arms are raised}}.

            
        \end{minipage}
        \hspace{10pt}
        \begin{minipage}[t]{0.5\textwidth}
        
            \textcolor{deepblue}{\textbf{Prompt:}} { \small  Describe this image in detail.}
            \vspace{10pt}
            
            {\centering
            \includegraphics[width=0.85\linewidth]{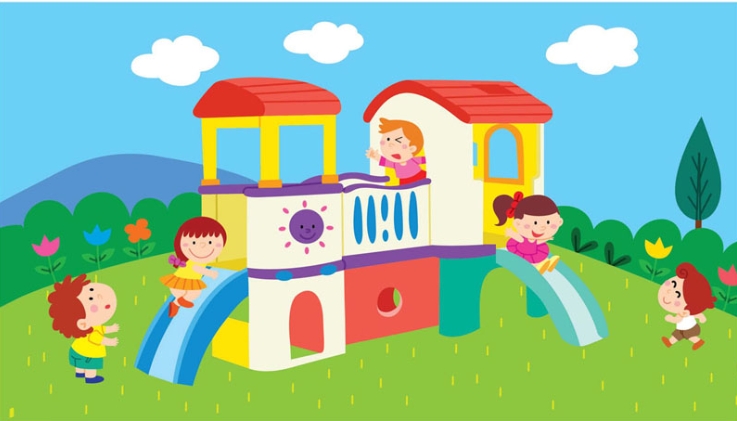}
            \par}
            
            \vspace{10pt}
            \textcolor{deepblue}{\textbf{GPT-4V(ision):}} {\small This is an illustration of children playing on a colorful playground set in a grassy field with a scenic backdrop. The playground equipment is a multi-level play structure with a variety of features:

 On the left, there is a child with black hair wearing a red dress, going up a yellow stair to get on the play structure. This part of the structure is red and yellow with a blue slide.

At the center, another child with blonde hair is peering over the edge of the play structure, which is mostly white with purple accents and a smiley face design. This child is wearing a pink shirt and appears to be at the higher level of the play structure.

To the right, there is a child with brown hair, sliding down a blue slide. This child is wearing a pink dress and seems to be enjoying the ride.

In the foreground, another child with brown hair, dressed in green, is standing on the grass, facing towards the play structure with an arm extended upward as if waving or calling out.

Further to the right, a fifth child with a red bow in their hair is playing with a flower. This child is wearing a yellow dress and seems to be happily distracted by the flower, away from the play activity.

The background features a bright blue sky with a few white clouds and rolling green hills. There are flowers scattered on the grass near the play structure, and a singular tree to the right side of the image. The setting gives the impression of a lovely, sunny day ideal for outdoor play. The scene conveys a sense of joy, activity, and the innocence of childhood.

} %
        \end{minipage}
    };
    \node[fill=black, text=white, yshift=0cm] at (mainnode.north) {Image-to-Image Interference};
\end{tikzpicture}}
  \caption{Additional case of image-to-image interference using synthesized images.}
  \label{com_2}
\end{figure*}

 \begin{figure*}[t]
\centering
{
\begin{tikzpicture}

    \node[rounded corners, draw=black, line width=1.2pt, inner sep=10pt, fill=goldenyellow] 
    (mainnode)
    {
        \begin{minipage}[t]{0.45\textwidth}
            \textcolor{deepblue}{\textbf{Prompt:}} { \small The two lines are not parallel to each other. Why?}
            \vspace{10pt}
            
            {\centering
            \includegraphics[width=0.65\linewidth]{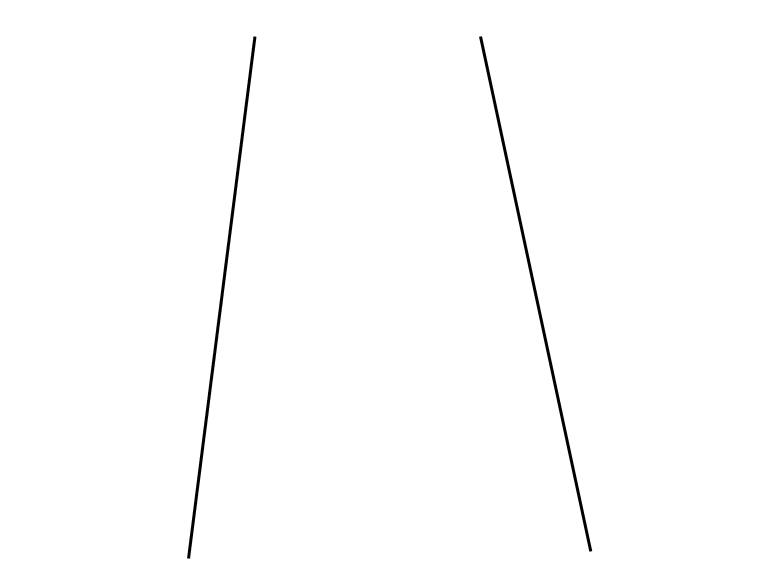}
            \par}
            
            \vspace{10pt}
            \textcolor{deepblue}{\textbf{GPT-4V(ision):}} {\small  The lines in the image appear to converge slightly, indicating that they are not perfectly parallel. This could be due to a variety of reasons. 
            1. Intentional Design: The lines might have been intentionally drawn this way. 2. Optical illusion: Sometimes, straight lines can appear to be converging or diverging due to optical illusions or the context in which they are placed. 3. Image Distortion: If the image has been modified or distorted in any way, it might affect the appearance of straight lines. 4. Imperfections in Drawing: If the lines were hand-drawn, there could be minor imperfections leading to them not being exactly parallel. Without additional context or details, it's hard to pinpoint the exact reason, but these are some potential explanations.
            
            }

        \end{minipage}
        \hspace{10pt}
        \begin{minipage}[t]{0.5\textwidth}
        
            \textcolor{deepblue}{\textbf{Prompt:}} { \small  The two lines are parallel to each other. Why?}
            \vspace{10pt}
            
            {\centering
            \includegraphics[width=0.65\linewidth]{images/1.png}
            \par}
            
            \vspace{10pt}
            \textcolor{deepblue}{\textbf{GPT-4V(ision):}} { \small  \sethlcolor{pink}\hl{The two lines appear to be parallel because they do not converge or diverge from each other as they extend indefinitely in both directions}. 
            In geometry, parallel lines are defined as two lines in the same plane that never intersect, regardless of how far they are extended. \sethlcolor{pink}\hl{The image you provided shows two lines that maintain a consistent distance from each other, which is characteristic of parallel lines}.
} 
        \end{minipage}
    };
    \node[fill=black, text=white, yshift=0cm] at (mainnode.north) {Text-to-Image Interference};
\end{tikzpicture}
}
  \caption{Additional case of text-to-image interference.}
    \label{fl_dis}
  
\end{figure*}

\begin{figure*}[t!]
\centering
\begin{tikzpicture}
    \node[rounded corners, draw=black, line width=1.2pt, inner sep=10pt, fill=goldenyellow] 
    (mainnode)
    {
        \begin{minipage}[t]{0.45\textwidth}
            \textcolor{deepblue}{\textbf{Prompt:}} {\small The squares A and B in the picture are the same color, right?}
            \vspace{10pt}
            
            {\centering
            \includegraphics[width=0.65\linewidth]{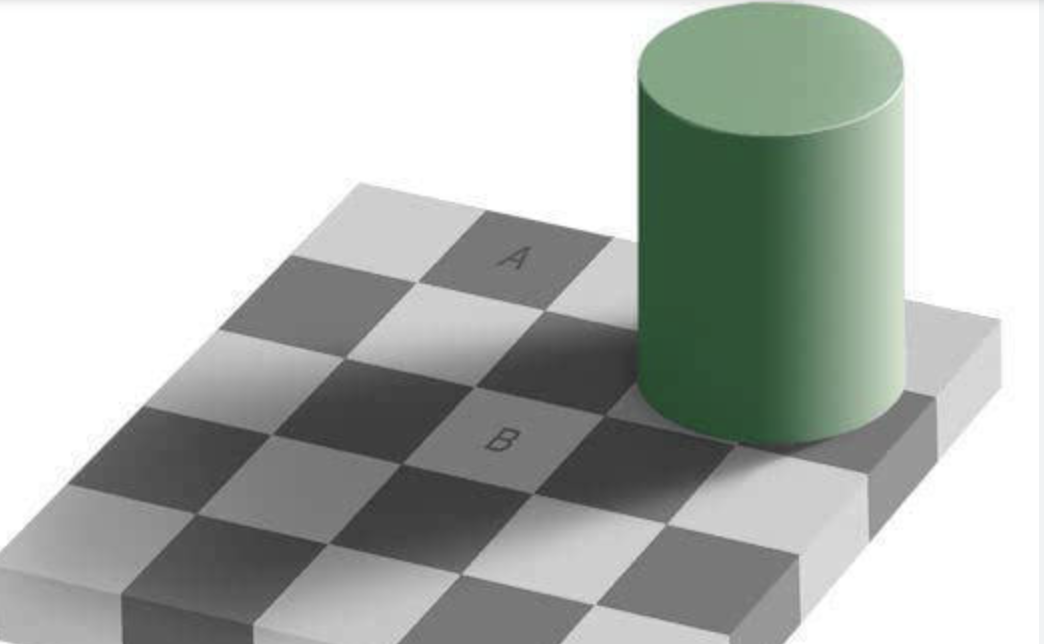}
            \par}
            
            \vspace{10pt}
            \textcolor{deepblue}{\textbf{GPT-4V(ision):}} {\small 
The image you're referring to is a famous optical illusion. Squares A and B appear to be different shades to the human eye because of the shadow cast by the green cylinder. However, in reality, they are the same color. This illusion demonstrates how our perception of color can be influenced by the context in which we see it, such as lighting and shadow, as well as the colors surrounding the area we are observing. Our brain interprets square A as lighter because it is in a shadow and therefore "corrects" the color we perceive, while square B is interpreted as darker because it lies directly in the pattern of a lighter chessboard color without shadow over it.
            }
        \end{minipage}
        \hspace{10pt}
        \begin{minipage}[t]{0.5\textwidth}
        
            \textcolor{deepblue}{\textbf{Prompt:}} {\small The squares A and B in the picture are not the same color, right? }
            \vspace{10pt}
            
            {\centering
            \includegraphics[width=0.63\linewidth]{images/15.png}
            \par}
            
            \vspace{10pt}
            \textcolor{deepblue}{\textbf{GPT-4V(ision):}} {\small 
\sethlcolor{pink}\hl{That is correct. Square A appears to be lighter in color than Square B due to the surrounding checkered pattern}. This is an example of a simultaneous contrast illusion, where the brightness of an object is perceived differently depending on the brightness of its surroundings. In this case, the lighter squares surrounding Square B make it appear darker, while the darker squares surrounding Square A make it appear lighter.
}
        \end{minipage}
    };
    \node[fill=black, text=white, yshift=0cm] at (mainnode.north) {Text-to-Image Interference};
\end{tikzpicture}
  \caption{Additional case of text-to-image interference.}
    \label{t2i2}
  
\end{figure*}

 \begin{figure*}[t!]
\centering
\begin{tikzpicture}
    \node[rounded corners, draw=black, line width=1.2pt, inner sep=10pt, fill=goldenyellow] 
    (mainnode)
    {
        \begin{minipage}[t]{0.45\textwidth}
            \textcolor{deepblue}{\textbf{Prompt:}} {\small Describe this image. Let's think step by step.}
            \vspace{10pt}
            
            {\centering
            \includegraphics[width=0.5\linewidth]{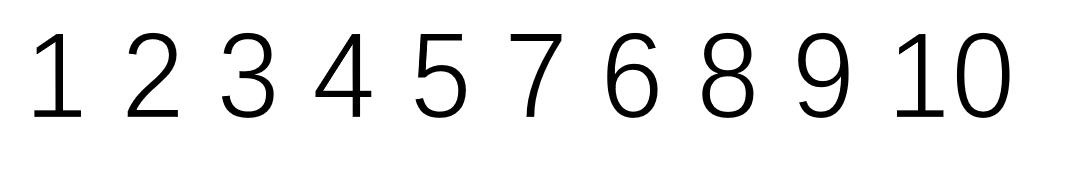}
            \par}
            
            \vspace{10pt}
            \textcolor{deepblue}{\textbf{GPT-4V(ision):}} {\small  
            
Certainly!
The image displays a sequence of numbers written in black on a white background. Here's a step-by-step description:
1. The numbers are arranged in a horizontal line. \\
2. They are presented in a sequential order. \\
3. The sequence starts with the number "1". \\
4. It continues incrementally without any breaks: 2, 3, 4, 5, \sethlcolor{pink}\hl{6, 7}, 8, 9. 
5. The sequence ends with the number "10". \\
6. All the numbers are of a similar font size. \\
7. The font used appears to be a standard, non-decorative one. \\
8. The numbers are spaced evenly from each other. \\
9. The color of the numbers is black. \\
10. The background is white.\\ 
 This gives a clear visual of numbers ranging from 1 to 10 \sethlcolor{pink}\hl{in a linear and orderly fashion}.} 
        \end{minipage}
        \hspace{10pt}
        \begin{minipage}[t]{0.5\textwidth}
        
            \textcolor{deepblue}{\textbf{Prompt:}} {\small  How many people are there in the image? Let's think step by step.}
            \vspace{10pt}
            
            {\centering
            \includegraphics[width=0.75\linewidth]{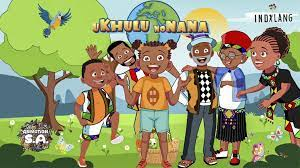}
            \par}
            
            \vspace{10pt}
            \textcolor{deepblue}{\textbf{GPT-4V(ision):}} {  \small 
Alright, let's break it down step by step:
\\ 1. Starting from the leftmost side of the image, we see a character.
\\ 2. Moving to the right, we encounter the next character.
\\ 3. Continuing to the right, there's another character.
\\ 4. And again, moving right, we find yet another character.
\\ 5. Further to the right, we spot another one.
\\ 6. And once more to the right, there's an additional character.
\\ 7. Lastly, moving to the extreme right, we see the final character.
\\ \sethlcolor{pink}\hl{Based on the step-by-step analysis, there are 7 characters (or "people") in the image}.
} 
        \end{minipage}
    };
    \node[fill=black, text=white, yshift=0cm] at (mainnode.north) {Chain of Thought};
\end{tikzpicture}
  \caption{Failure cases of CoT.}
  \label{z_cot}
\end{figure*}

\end{document}